\documentclass[conference]{IEEEtran}
\IEEEoverridecommandlockouts  
\usepackage{amsmath}
\usepackage{multirow}
\usepackage{algorithm}
\usepackage{algorithmic}
\usepackage{amsfonts}
\usepackage{subfig}
\usepackage{graphics}
\usepackage{graphicx}
\usepackage{amssymb}
\usepackage{caption}
\usepackage{array}
\usepackage{amsmath}
\usepackage{hyperref}
\usepackage{mathtools}
\usepackage{cite}
\usepackage{pgfplots}
\usepackage{textcomp}
\usepackage{arydshln}
\usepackage{soul}
\hypersetup{bookmarks=false, hidelinks=true}
\begin{document}
\title{\huge Search-Based Online Trajectory Planning for \\ Car-like Robots in Highly Dynamic Environments }

\author{\authorblockN{Jiahui Lin$^{1}$, Tong Zhou$^{1}$, Delong Zhu$^{1*}$, Jianbang Liu$^{1}$, and Max Q.-H. Meng$^{1,2*}$}
\thanks{$^*$ The corresponding author of this paper.}	
\thanks{$^1$ The authors are with the Department of Electronic Engineering, The Chinese University of Hong Kong, Shatin, N.T., Hong Kong SAR, China. \textit{email: zhudelong@link.cuhk.edu.hk, jiahuilin@cuhk.edu.hk, tzhou@ee.cuhk.edu.hk} }
\thanks{$^2$ Max Q.-H. Meng is with the Department of Electronic and Electrical Engineering, Southern University of Science and Technology, Shenzhen, China, and also with the Shenzhen Research Institute of the Chinese University of Hong Kong, Shenzhen, China, on leave from the Department of Electronic Engineering of the Chinese University of Hong Kong, Hong Kong (e-mail: max.meng@ieee.org). This project is partially supported by the Hong Kong RGC GRF grants \#14200618 and Hong Kong ITC ITSP Tier 2 grant \#ITS/105/18FP awarded to Max Q.-H. Meng.}	
}

\maketitle
\begin{abstract}
This paper presents a search-based partial motion planner to generate dynamically feasible trajectories for car-like robots in highly dynamic environments. The planner searches for smooth, safe, and near-time-optimal trajectories by exploring a state graph built on motion primitives, which are generated by discretizing the time dimension and the control space. To enable fast online planning, we first propose an efficient path searching algorithm based on the aggregation and pruning of motion primitives. We then propose a fast collision checking algorithm that takes into account the motions of moving obstacles. The algorithm linearizes relative motions between the robot and obstacles and then checks collisions by comparing a point-line distance. Benefiting from the fast searching and collision checking algorithms, the planner can effectively and safely explore the state-time space to generate near-time-optimal solutions. The results through extensive experiments show that the proposed method can generate feasible trajectories within milliseconds while maintaining a higher success rate than up-to-date methods, which significantly demonstrates its advantages.       
\end{abstract}

\section{Introduction}
Online trajectory planning in dynamic environments has a wide application in autonomous robotic systems \cite{hawkeye-zhu, drl-zhu, Tingguang2019Learning, chaoqunsrm}. However, it is a very challenging task due to unpredictable future motions of moving obstacles. The nonholonomic model and time-dependent collision checking impose highly nonlinear constraints on the planning task, making the problem even more intractable. The optimal solution is typically unachievable, and the planning completeness also cannot be ensured \cite{survey2020}. As a result, existing studies mostly focus on finding near-optimal partial solutions within a specific planning horizon to approximate the global solution with the help of a fast re-planning process.   

Partial Motion Planning (PMP) in dynamic environments is first proposed by Petti \textit{et al.} \cite{ics-safe-planning05} under the framework of Rapidly Exploring Random Tree (RRT). Since in dynamic environments, the randomized sampling approach, e.g., RRT, takes a long time to converge, the authors allow a time-bounded partial solution with its terminal state ICS-free \cite{ics-theory04}. In this way, the planning and execution procedures can be accurately scheduled to achieve fast re-planning. Based on this idea, more sampling-based methods, e.g., \cite{drt-17}, are proposed to improve the optimality of partial trajectories. The idea of PMP is also adopted by some searching-based algorithms \cite{dwa07, lattice2, lattice3, lattice4}, which combines a long-horizon path searcher for improving the global optimality and a local collision avoidance strategy to ensure safety. The major deficiency of these methods lies in planning inefficiency. There typically exist a large number of candidate paths in a time-involved planning problem, and the collision checking at each searching step also needs a complicated procedure to cope with the motions of obstacles, which significantly limits the planning speed. Meanwhile, the differential constraints will reduce the volume of admissible state space, making the problem even more difficult.   
      
To address these problems, we propose an online partial motion planner to generate dynamically feasible trajectories in highly dynamic environments. We first discretize the time dimension and control space of the robot to generate motion primitives, which are short trajectories within a specified duration. Continuous expansion of the primitives will generate a primitive tree that covers the state space. Direct searching on the primitive tree is a time-consuming process. We thus also discretize the state space into grids. Primitives that lie in the same grid are aggregated and then pruned. The pruning operation is carefully designed, which can significantly reduce the searching scale while maintaining the smoothness at pruning points. Furthermore, we propose an efficient collision checking algorithm that takes into account the motions of obstacles (modeled by velocity extrapolation method \cite{survey2020}). The algorithm performs velocity planning on the entire motion primitive and can provide a strict safety guarantee under the velocity extrapolation model.

The contributions of this work are as follows:
\begin{itemize}
	\item We propose an efficient online graph searching algorithm based on motion primitives. By leveraging node aggregation and pruning, the algorithm can efficiently explore the state space and generate near-time-optimal solutions.  
	\item We propose a fast collision checking algorithm based on the linearization of relative motions between the robot and moving obstacles, which significantly improves the safety of the generated trajectory.   
	\item Based on the searching and collision checking algorithms, we implement an online trajectory planning framework with very high planning frequency and success rates. The code will also be available soon.
\end{itemize}

\section{Related work}
\textbf{Motion primitive} is a frequently used method by the planning community. LaValle \textit{et al.} \cite{primitive1} propose a node expansion algorithm using motion primitives to enable kinodynamic planning with RRT. The idea is to approach the target state with motion primitives to make the trajectory satisfy kinodynamic constraints. However, the primitives used in \cite{primitive1} are generated by sampling the control space rather than solving the two-point Boundary Value Problem (BVP), hence the continuousness of the trajectory is not ensured.  This problem is then solved by Luigi \textit{et al.} \cite{primitive3} using an efficient BVP solver.  
Liu \textit{et al.} \cite{primitive4} combine the graph searching method with motion primitives to approximate the optimal control problem for quadrotors, achieving aggressive flight in SE(3) space \cite{primitive5}. The above methods are all designed for static environments but share a similar idea with ours. 

\textbf{State lattice} is another discretization method that transforms the continuous state space into searching graphs.
The state lattice is introduced by \cite{lattice1} for motion planning in static environments. The edges that connect adjacent states in state lattice are another type of motion primitive. Matthew \textit{et al.} \cite{lattice2} extend the state lattice with a time dimension, which allows the search algorithm to explore both spatial and temporal dimensions efficiently. 
Kushleyev \textit{et al.} \cite{lattice3} propose a time-bounded lattice for planning in dynamic environments. The A* is used to search on a pre-computed lattice. However, the collision checking in this work is conducted on the discretized waypoints, thus cannot provide a safety guarantee. 

\textbf{Trajectory library} can be regarded as the third type of motion primitive. The idea is to build a prior road map that respects motion constraints for the target environment, and then select the optimal path based on an online collision checking process.
Zhang \textit{et al.} \cite{offline2} adopt the BIT* method \cite{offline3} to pre-build a prior map for the environment, based on which an online collision checking process is conducted to select the best trajectory. In this way, the computation cost is significantly reduced. Some other studies that make use of prior maps include Vector Field based method \cite{offline5} and Voronoi Random Fields based method \cite{ref15}. Both methods aim to accelerate the online processing by downloading part of the computation to an offline process. The trajectory library based method is capable of dealing with dynamic environments to some extent. For highly dynamic environments, the number of trajectories that need collision checking will be intractable.

\section{Problem Definition}
The planning problem in this work is defined as a tuple $\mathcal{<R, M, B, J>}$, which denotes the motion model, the map representation, the boundary value, and the planning objective.

The motion model $\mathcal{R}$ of a mobile robot is defined by its state function with differential constraints,
\begin{equation}
 \begin{aligned}
	\dot{\mathbf{x}}&=f(\mathbf{x}, \mathbf{u}), \\
	h(\mathbf{x}, \mathbf{u})&=0,\; h(\mathbf{x}, \mathbf{u}) \leq 0,
	\label{motion-c}
 \end{aligned}
\end{equation}
where $\mathbf{x}$ and $\mathbf{u}$ are robot state and control input, respectively. 

The map representation $\mathcal{M}$ defines a model for the environment, which separates the free space $\mathcal{M}_{f}$ from static obstacles. In dynamic environments, due to the presence of moving obstacles, $\mathcal{O}_i(t), i\in \{1, \cdots, M\}$, the free space is further restricted, which imposes safety constraints on the planning,    
\begin{equation}
	\mathbf{x}(t) \notin \mathcal{M}_{f}\cap \mathcal{O}_{i}(t), \; \forall i \in\{1, \ldots, M\}, \forall t \in [t_0, t_0+T],
	\label{env-c}
\end{equation}
where $t_0$ is the start time of collision checking, and $T$ is the planning horizon. One of the key challenges of planning in dynamic environments lies in the calculation of such constraints.

The boundary value $\mathcal{B}$ specifies a planning start $\mathbf{x}_0$, a goal set $\mathcal{X}_g$, and an intermediate point $\mathbf{x}_T \not \in \mathcal{X}_{ICS}$. For planning without moving obstacles, the intermediate point is typically not necessary, since any point in a successfully planned path is ensured to be collision-free during execution. However, this condition does not hold, when there exist moving obstacles, since their motions are unpredictable. We thus need to specify a local ICS-free goal to ensure the robot can safely navigate there before the next trajectory is available. This imposes the boundary value constraints on the planning problem,
\begin{equation}
	\mathbf{x}(t_0) = \mathbf{x}_0, \; \mathbf{x}(t_0 + T_g) \in \mathcal{X}_g, \; \mathbf{x}(t_0 + T) \not \in \mathcal{X}_{ICS}.
	\label{bound-c}
\end{equation}  

The planning objective in this work is to generate a trajectory that is smooth, collision-free, respects the motion constraints, and has minimum execution time. The smoothness is defined as the squared L2-norm of the control efforts, 
\begin{equation}
J =\int_{0}^{T_g}\|\mathbf{u}(t)\|^{2} d t.
\end{equation}
The less control efforts are applied, the less state changes are obtained, and thus the generated trajectory is smoother.

Based on the above definition, we formulate the dynamical motion planning into an optimization problem, 
\begin{equation}
\begin{aligned}
	&\min _{\mathbf{x}(t), \mathbf{u}(t), T_g} J+ \beta T_g \\
	\dot{\mathbf{x}}=f(&\mathbf{x}, \mathbf{u}), \; h(\mathbf{x}, \mathbf{u})=0, \; h(\mathbf{x}, \mathbf{u}) \leq 0,\\
	\mathbf{x}(t) \notin & \;\mathcal{M}_{f}\cap \mathcal{O}(t), \;  t \in [t_0, t_0+T], \\
	\mathbf{x}(t_0) = \mathbf{x}_0, \; &\mathbf{x}(t_0+T_g) \in \mathcal{X}_g, \; \mathbf{x}(t_0 + T) \not \in \mathcal{X}_{ICS}.
	\label{opt}
\end{aligned}
\end{equation}
The solution for this problem defines a partial trajectory within the planning horizon $T$. For the remaining part that lies in $(T, T_g]$, the safety constraint is not required to be strictly satisfied. This is different from the conventional motion planning problem and is referred to as the PMP. 
Optimizing Eq. \eqref{opt} in continuous space is intractable. We thus approximate the solution with motion primitives at the expense of optimality.

\section{Graph Searching with Motion Primitives}

\subsection{Robot Motion Model}
There are many types of motion models for wheeled robots, including holonomic and nonholonomic ones. In this work, as shown in Fig. \ref{rgb_pc}, we adopt the canonical simplified car kinematics to model the robot,
\begin{equation}
	\dot{\mathbf{x}}=\left[\begin{array}{c}
	\dot{\phi} \\
	\dot{x} \\
	\dot{y}
	\end{array}\right]=B(\mathbf{x}) \mathbf{u}=\left[\begin{array}{cc}
	0 & 1 \\
	\cos \phi & 0 \\
	\sin \phi & 0
	\end{array}\right]\left[\begin{array}{c}
	v \\
	\omega
	\end{array}\right],
	\label{motion-model}
\end{equation}
with control transformations
\begin{equation}
	v = v_0+a*t, \; \omega = (\tan\psi) / \ell * v,
	\label{control-trans}
\end{equation}
where $\mathbf{x}\small{=}(x, y, \phi) \in \mathcal{X}$ defines the robot state, which includes the heading direction $\phi$ and robot location $(x, y)$, and $\psi$ is the steering angel. The transformed control input $(v, \omega)$ includes the forward speed and rotation rate of the robot. A feasible control set is visualized in Fig. \ref{rgb_img}, where $v$ and $\omega$ are linearly related, as indicated by Eq. \eqref{control-trans}. The slope rate of the lines in the bowtie control set is determined by the maximum steering angel $\psi$ (or the minimum turning radius) of the robot. 

In practice, a more frequently-used control input is the steering angle and linear acceleration, i.e., $\mathbf{u} =(\psi, a)$. Each $\psi$ corresponds to a line in Fig. \ref{rgb_img} with its slope rate defined by $(\tan\psi)/\ell$. If we denote the current velocity of the robot by $v_c$ and the steering angle is not changed, the subsequent velocities accelerated by $a$ will fall on this line. 
%
To make the generated trajectory dynamically feasible, we need to make sure the control inputs along the trajectory always lie in the control set, i.e., $\mathbf{u}(t) \in \mathcal{U}$.

\subsection{Motion Primitives}  
\begin{figure}[t]
	\centering
	\subfloat[The simplified car kinematics.]{		
		\includegraphics[width=0.44\linewidth]{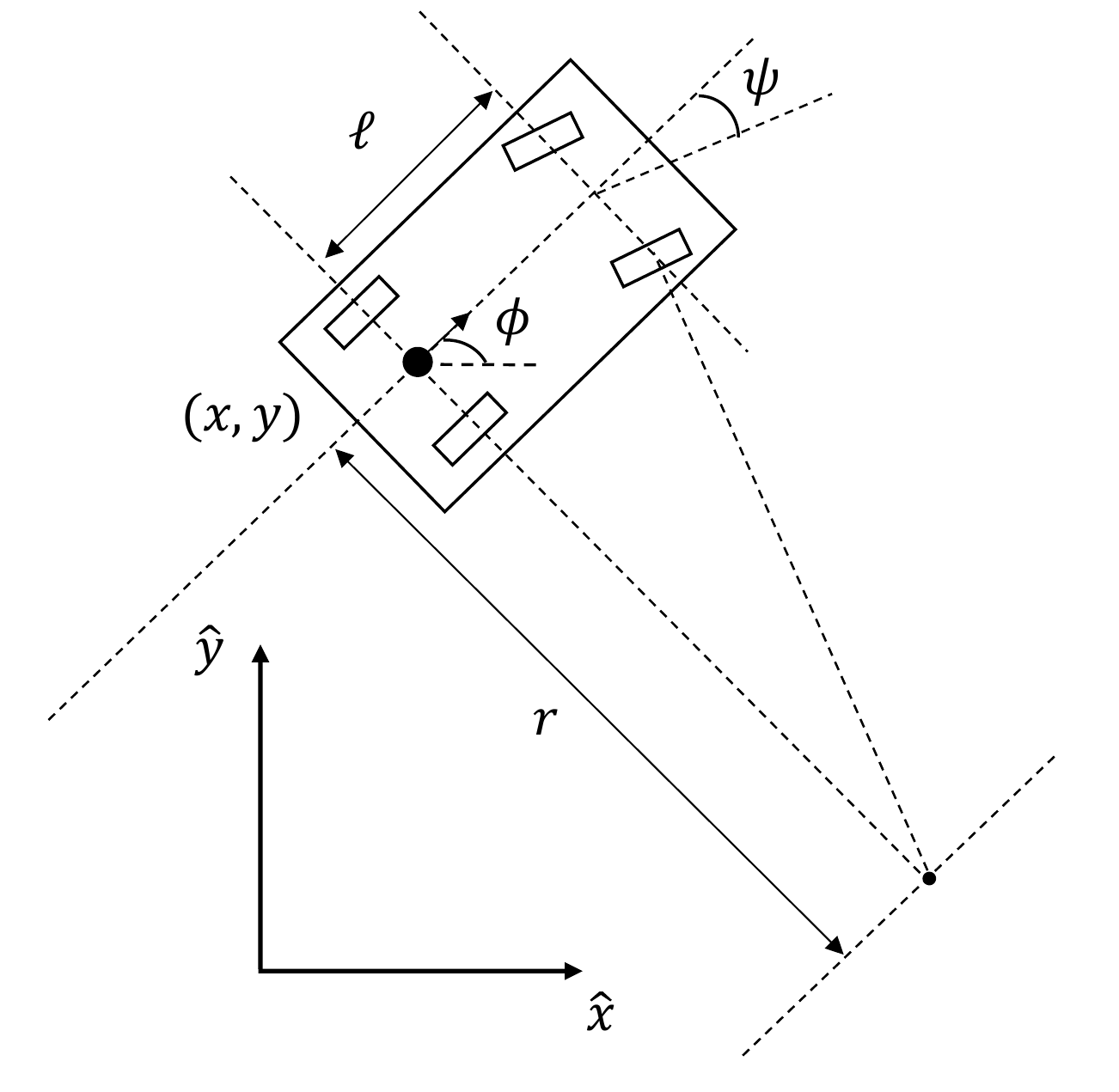}
		\label{rgb_pc}	
	}
	\subfloat[The $(v, w)$ control set.]{		
		\includegraphics[width=0.44\linewidth]{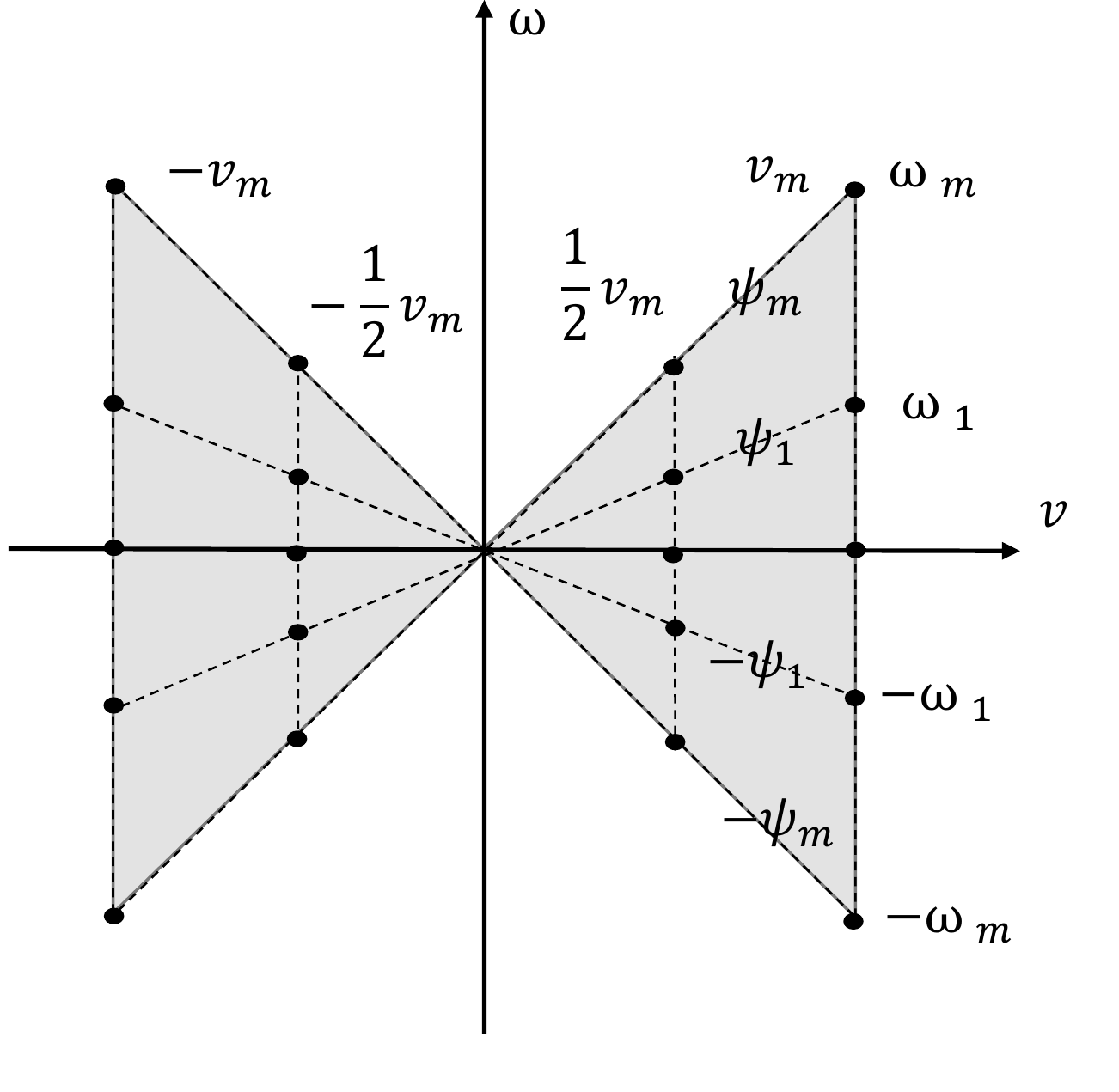}	
		\label{rgb_img}
	}
	\caption{The motion and control models for the wheeled robot.}
	\label{fig-models}
	\vspace{-4mm}
\end{figure} 

Motion primitive $\boldsymbol{\xi}_k$ is a short trajectory within duration $\tau$, which can bring the robot from the current state $\mathbf{x}_k$ to a new state $\mathbf{x}_{k+1}$ while respecting the motion constraints. 
Denote by $\mathcal{U}_{N}=\left\{ \mathbf{u}_{1}, \ldots, \mathbf{u}_{N}\right\} \subset \mathcal{U}$ the discretized control inputs, the motion primitives for $t \in[0, \tau]$ that apply $\mathbf{u}_{n} \in \mathcal{U}_{N}$ are generated as follows,  
\begin{equation}
\begin{aligned}
\phi(t)& =\phi_{k}+(v_ct+\frac{1}{2}a_n t^{2})(\tan\psi_n)/\ell,\\
x(t)&=x_k+\sin(\phi(t))\,\ell/(\tan\psi_n),\\
y(t)&=y_k-\cos(\phi(t))\,\ell/(\tan\psi_n).\\
\end{aligned}
\label{eq-solution}
\end{equation}
where $(\tan\psi_m)/\ell = 1/r$ and $r$ is the turning radius (Fig. \ref{rgb_pc}) of the primitive.
It is worth noting that the linear velocity sometimes may reach its maximum before $\tau$ is used up. In this case, the remaining of the motion primitive is generated with the maximum velocity.

\begin{figure}[t]
	\centering
	\subfloat[One-layer motion primitives.]{		
		\includegraphics[width=0.48\linewidth]{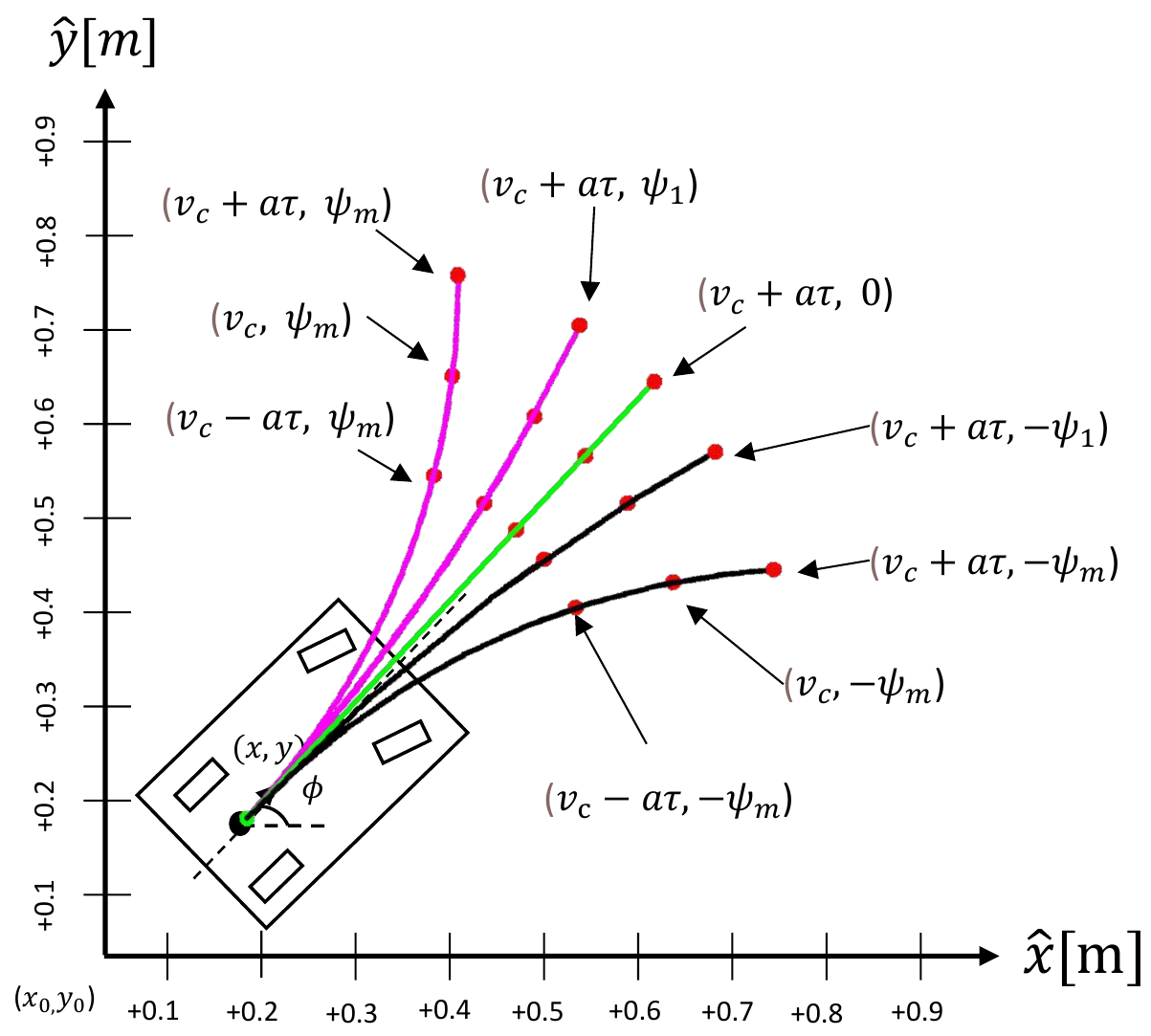}
		\label{fig-primitive}	
	}
	\subfloat[Motion primitive tree.]{		
		\includegraphics[width=0.48\linewidth]{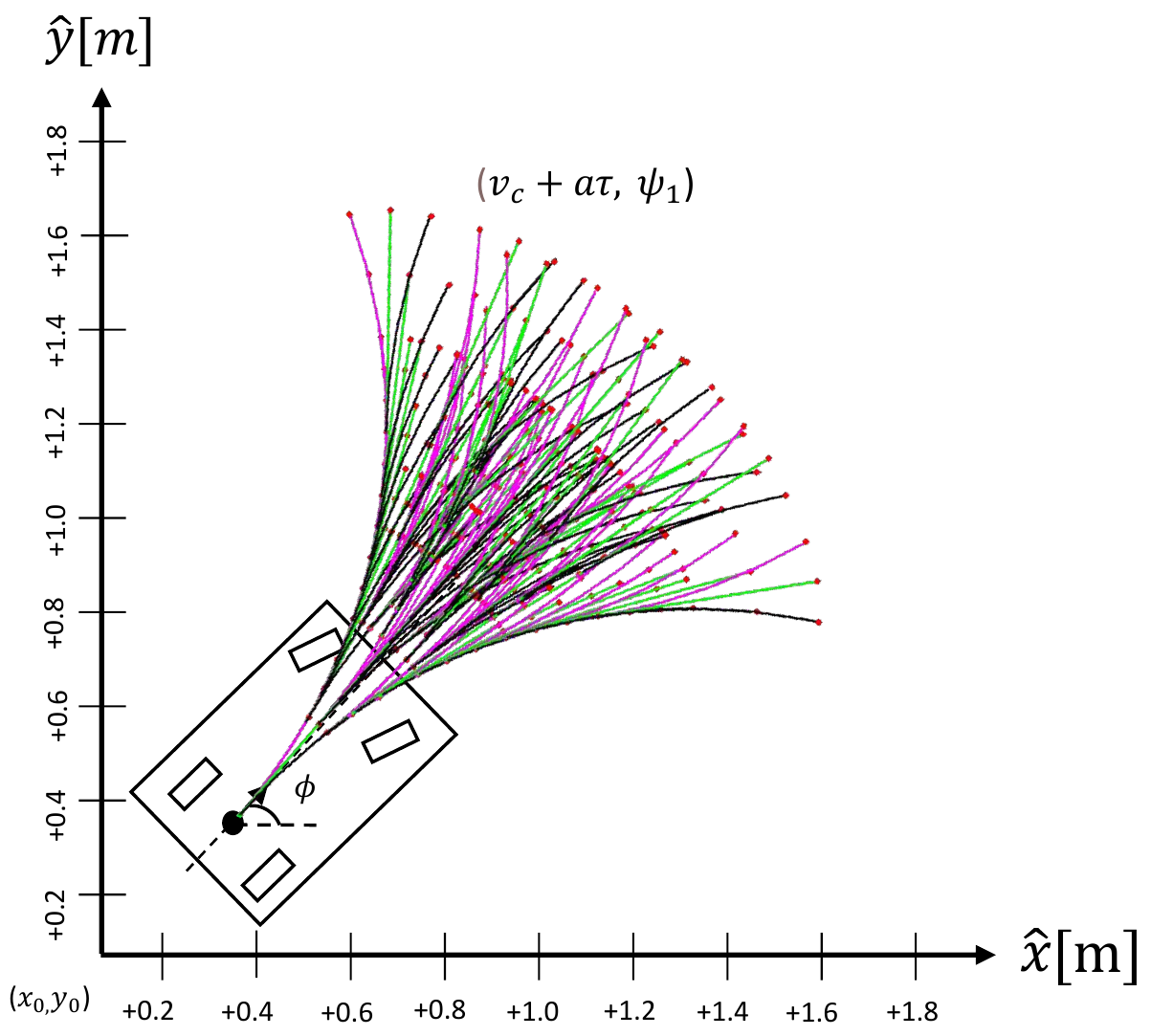}	
		\label{fig-lattice}
	}
	\caption{The generation of motion primitives based on a car model.}
	\label{fig-primitives}
\end{figure}

We visualize the one-layer motion primitives starting at $\mathbf{x}_k$ in Fig. \ref{fig-primitive}. Each arch is specified by a sampled steering angel, and each red point in the arch is a terminal state of the motion primitive generated with a sampled linear acceleration. The transformed control inputs along the motion primitive are constrained on a line segment within the control set, as shown in Fig. \ref{rgb_img}. Starting from the current state, we can forward sample the control inputs and continuously grow the motion primitives until the considered state space is sufficiently expanded. We demonstrate such an expansion process in Fig. \ref{fig-lattice}, which is essentially a tree structure and named primitive tree. 

\begin{algorithm}[tb]
	\caption{GeneratePrimitives($\mathbf{s}_k$, $\mathcal{X}_g$, $\mathcal{O}$)}
	\label{Generate-Succ}
	\begin{algorithmic}[1]
		\STATE \textsc{Succ} $\gets \emptyset$
		\FOR{$\mathbf{u}$ $\in \mathcal{U}_N$}
		\STATE $\boldsymbol{\xi}_k, \mathbf{s}_{k+1}$ $\gets$ \textsc{MotionPrimitive($\mathbf{s}_k, \mathbf{u}, \tau$)}
		\IF{\textsc{IsValid}$(\mathbf{s}_{k+1})$ $\land$ \textsc{CollisionFree}($\boldsymbol{\xi}_k, \mathcal{O}, \tau$) }
		\STATE  g($\mathbf{s}_{k+1}$) $\gets$ g($\mathbf{s}_k$) + \textsc{PiritiveCost($\boldsymbol{\xi}_k$)}
		\STATE  h($\mathbf{s}_{k+1}$) $\gets$ \textsc{Heuristic($\mathbf{s}_{k+1}, \mathcal{X}_g$)  }
		\STATE  f\,($\mathbf{s}_{k+1}$) $\gets$ g($\mathbf{s}_{k+1}$) + h($\mathbf{s}_{k+1}$)
		\STATE  $\rm\mathbf{s}_{k+1}.pre$ $\gets$ $\mathbf{s}_k$
		\STATE  \textsc{Insert}($\mathbf{s}_{k+1}$, \textsc{Succ})
		\ENDIF
		\ENDFOR
		\RETURN \textsc{Succ}
	\end{algorithmic}
\end{algorithm}
\subsection{Online Graph Searching}
Motion primitive $\boldsymbol{\xi}_k$ is uniquely determined by a tuple $(\mathbf{x}_k, \mathbf{u}_n, \tau)$, which is a sample of the solution space in Eq. \eqref{opt}, and the cost of each primitive is $\|\mathbf{u}(t)\|^{2}+\beta\tau$, where $\|\mathbf{u}(t)\|^{2}=p_{1}a^2 +p_{2} \psi^2$.
For the primitve with control $(a=0,\psi=0)$, $p_1=1,p_2=1$. Otherwise, $p_1$ and $p_2$ is set to $>1$. As a result, the planner prefers to approach the goal with maximum velocity in a straight line. 
By leveraging motion primitives, the PMP defined in Eq. \eqref{opt} is transformed into a combinatorial optimization problem over the discretized space,
\begin{equation}
\small
\begin{aligned}
\min _{\boldsymbol{\xi}_{0:K}, \; K} &\left(\sum_{k=0}^{K}\|\mathbf{u}_k\|^{2}+\beta (K+1)\right) \tau \\
\boldsymbol{\xi}_k((k+1)\tau) &= \boldsymbol{\xi}_{k+1}((k+1)\tau), \; \forall \boldsymbol{\xi}_i \in \Xi_i, \; \\
\boldsymbol{\xi}_k(t) \notin \mathcal{M}_{f}\cap \mathcal{O}(t), \; t &\in [k\tau,(k+1)\tau], \; \forall k  \in\{0, \ldots, \lceil T/\tau \rceil\}, \\
\boldsymbol{\xi}_0(0) = \mathbf{x}_0, \; &\boldsymbol{\xi}_K \cap \mathcal{X}_g  \not= \emptyset,\; \boldsymbol{\xi}_{\lceil T/\tau \rceil} \cap \mathcal{X}_{ICS}  = \emptyset,
\label{optd}
\end{aligned}
\end{equation}   
\begin{algorithm}[!htb]
	\caption{GraphSearching($\mathbf{s}_0$, $\mathcal{X}_g$, $\mathcal{O}$, $\tau$, $T$)}
	\label{astar}
	\begin{algorithmic}[1]
		\STATE \textsc{HMAP} $\gets \emptyset$, \textsc{OPEN} $\gets \emptyset$   \\
		\STATE $\rm\mathbf{s}_0.key \gets \textsc{GenKey}(\mathbf{s}_0)$, $\rm\mathbf{s}_0.idx \gets 0$
		\STATE $\rm\mathbf{s}_0.state \gets \mathbf{x}_0$, g($\mathbf{s}_0$) $\gets$ 0 \\
		\STATE \textsc{HMAP.Insert}($\mathbf{s}_0$), \textsc{OPEN.Insert}($\mathbf{s}_0$)
		\WHILE{\textsc{OPEN} is not empty}{
			\STATE $\mathbf{s}_k$ $\gets$ OPEN.\textsc{Pop()} \\
			\STATE $\mathbf{s}_k.\rm closed \gets \rm true$\\
			\STATE $\mathbf{s}_g$ $\gets$ \textsc{GoalReached}($\mathbf{s}, \mathcal{X}_g$) \\
			\STATE $\mathbf{s}_t$ $\,\gets$ \textsc{TimeReached}($\mathbf{s}, T$) \\
			\STATE /* \textit{the goal or planning horizon is reached} */
			\IF{$\mathbf{s}_g.\rm state \in \mathcal{X}_g $} 
			\RETURN \textsc{BackTrack}($\mathbf{s}_g$) \\
			\ELSIF{$\mathbf{s}_t.\rm state \not\in \mathcal{X}_{\textit{ICS}}$} 
			\RETURN \textsc{BackTrack}($\mathbf{s}_t$) \\
			\ENDIF
			\STATE /* \textit{generate successors using motion primitives} */  \\ 
			\STATE \textsc{Succ} $\gets$ GeneratePrimitives($\mathbf{s}_k$, $\mathcal{X}_g$, $\mathcal{O}$)  \\
			\FOR{$\mathbf{s}_{k+1} \in$ \textsc{Succ}}
			\STATE $\rm\mathbf{s}_{k+1}.key \gets \textsc{GenKey}(\mathbf{s}_{k+1})$, $\rm\mathbf{s}_{k+1}.idx$ $\gets 0$
			\STATE $\rm\mathbf{s}_{k+1}.state \gets \mathbf{x}_{k+1}$
			\STATE $\rm elements$ $\gets$ \textsc{HMAP.Query}($\rm\mathbf{s}_{k+1}.key$)
			\STATE /* \textit{already reached by other primitives} */  \\ 
			\IF{$\rm elements$ $\not =\emptyset$ }
			\STATE $n\, \gets$ \textsc{Length}($\rm elements$)\\
			\STATE $\mathbf{s^*} \gets$ $\rm elements[n-1]$\\
			\STATE $\rm case1 \gets \mathbf{s}^*.key=\mathbf{s}.key$ \\
			\STATE $\rm case2 \gets\mathbf{s^*}.pre = \mathbf{s}_{k+1}.pre$ \\
			\STATE $\rm case3 \gets \small{\sim}(case1 \,\vee\, case2)$\\
			\IF{($\rm case1 \,\vee\, case2$) $\wedge$ (h($\mathbf{s}_{k+1}$) $<$ h($\mathbf{s^*}$))}
			\STATE $\rm\mathbf{s}_{k+1}.idx$ $\gets n$\\
			\STATE \textsc{HMAP.Insert}($\mathbf{s}_{k+1}$), \textsc{OPEN.Insert}($\mathbf{s}_{k+1}$)
			\ENDIF
			\IF{$\rm case3 \; \wedge {s^*}$. closed$==$false $\wedge$ g($\mathbf{s}_{k+1}$) $<$ g($\mathbf{s^*}$)}
			\STATE $\rm\mathbf{s}_{k+1}.idx \gets \mathbf{s^*}.idx$\\
			\STATE  \textsc{HMAP.Replace}($\mathbf{s^*}, \mathbf{s}_{k+1}$)
			\ENDIF
			\STATE \hspace{-4mm}/* \textit{never reached by other primitives} */  \\ 
			\ELSE
			\STATE \textsc{HMAP.Insert}($\mathbf{s}_{k+1}$), \textsc{OPEN.Insert}($\mathbf{s}_{k+1}$)\\
			\ENDIF    
			\ENDFOR
		}
		\ENDWHILE
		\RETURN \textsc{Failure}  
	\end{algorithmic} 
\end{algorithm}
\begin{figure}[t]
	\vspace{-4mm}
	\centering
	\subfloat[case1: $\mathbf{s}_k$ and $\mathbf{s}_{k+1}$ are in the same grid cell, and $\mathbf{s}^*$ is the parent or sibling of $\mathbf{s}_{k+1}$.]{		
		\includegraphics[width=0.39\linewidth]{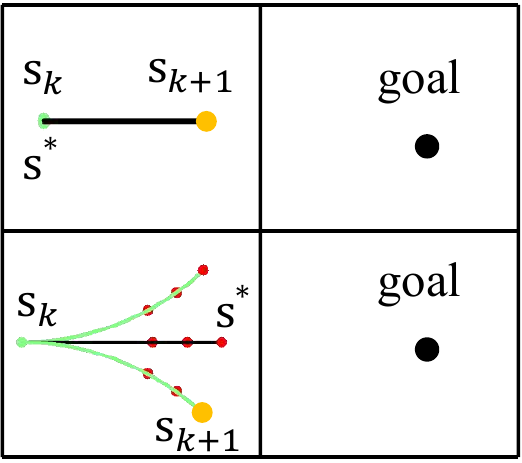}
	}\hspace{1mm}
	\subfloat[case2: $\mathbf{s}_k$ and $\mathbf{s}_{k+1}$ connect two grid cells, and $\mathbf{s}^*$ is the sibling of $\mathbf{s}_{k+1}$.]{		
		\includegraphics[width=0.39\linewidth]{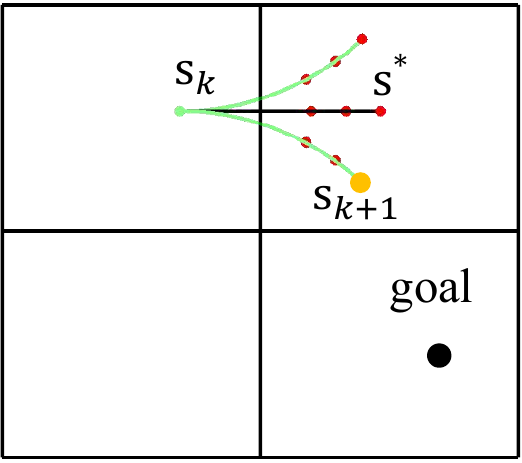}
	}
	
	\subfloat[case3: $\mathbf{s}_k$ and $\mathbf{s}_{k+1}$ connect two grid cells, and $\mathbf{s}^*$ is not the parent or sibling of $\mathbf{s}_{k+1}$. ]{		
		\begin{minipage}[b]{0.82\linewidth}
			\includegraphics[width=1.0\linewidth]{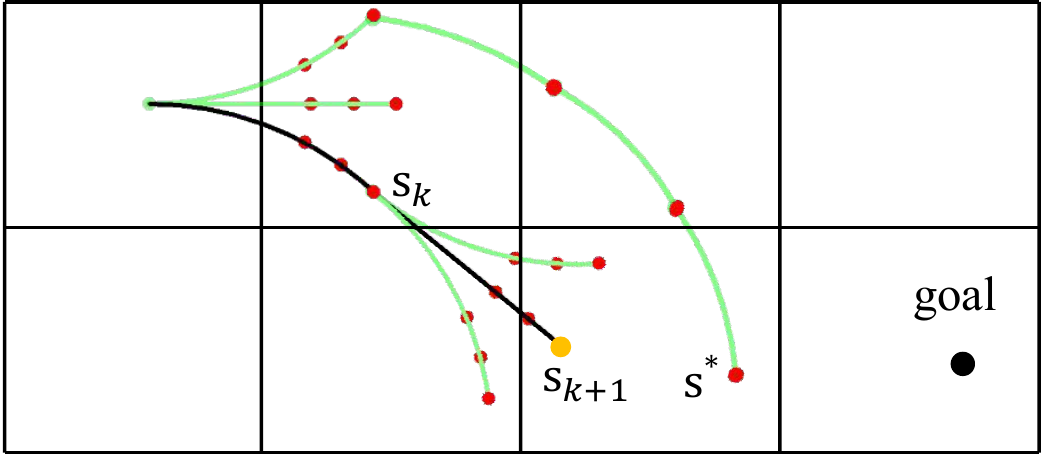}\vspace{8pt}
		\end{minipage}	
	}
	\caption{Illustration of the three cases of node expansion. $\mathbf{s}^{*}$ is the optimal representation of the current grid cell. $\mathbf{s}_{k}$ is the parent node and $\mathbf{s}_{k+1}$ is one of the child nodes being evaluated. The yellow point is the newly selected optimal representation of the current grid cell.}
	\label{special-case}
	\vspace{-4mm}
\end{figure}
where $\Xi_i$ is the set of motion primitives generated for time duration $[i\tau, (i+1)\tau]$ with a depth of $i$ in the primitive tree. Based on such a discretization scheme, the planning problem becomes finding a sequence of collision-free motion primitives that connect the start and goal within the planning horizon while maintaining an ICS for the local goal.

\begin{figure*}[ht]
	\centering
	\subfloat[Illustration of the linearized collision checking.]{		
		\includegraphics[width=0.43\linewidth]{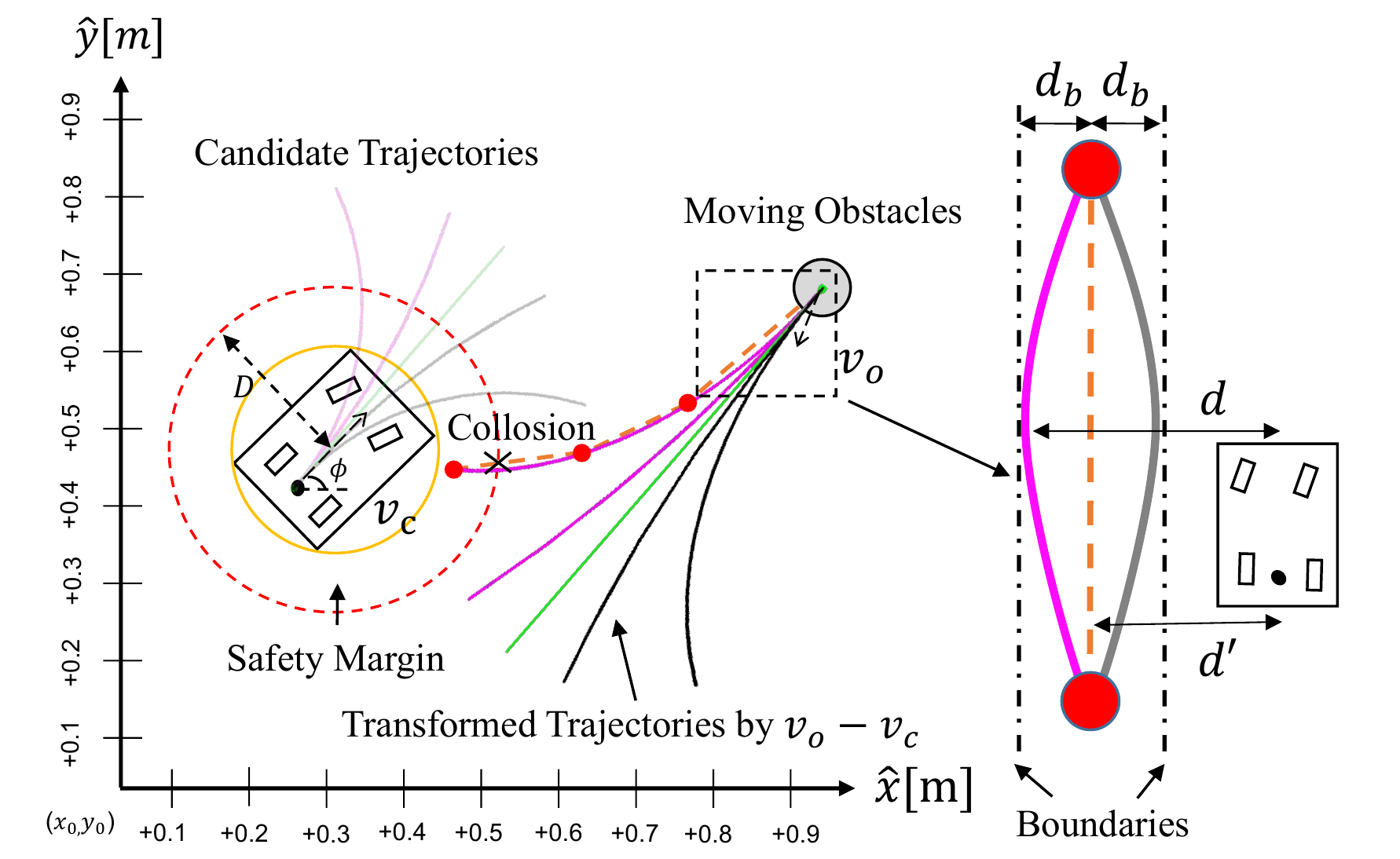}
		\label{fig-collision}	
	}
	\subfloat[Illustration of the process of ICS checking.]{		
		\includegraphics[width=0.53\linewidth]{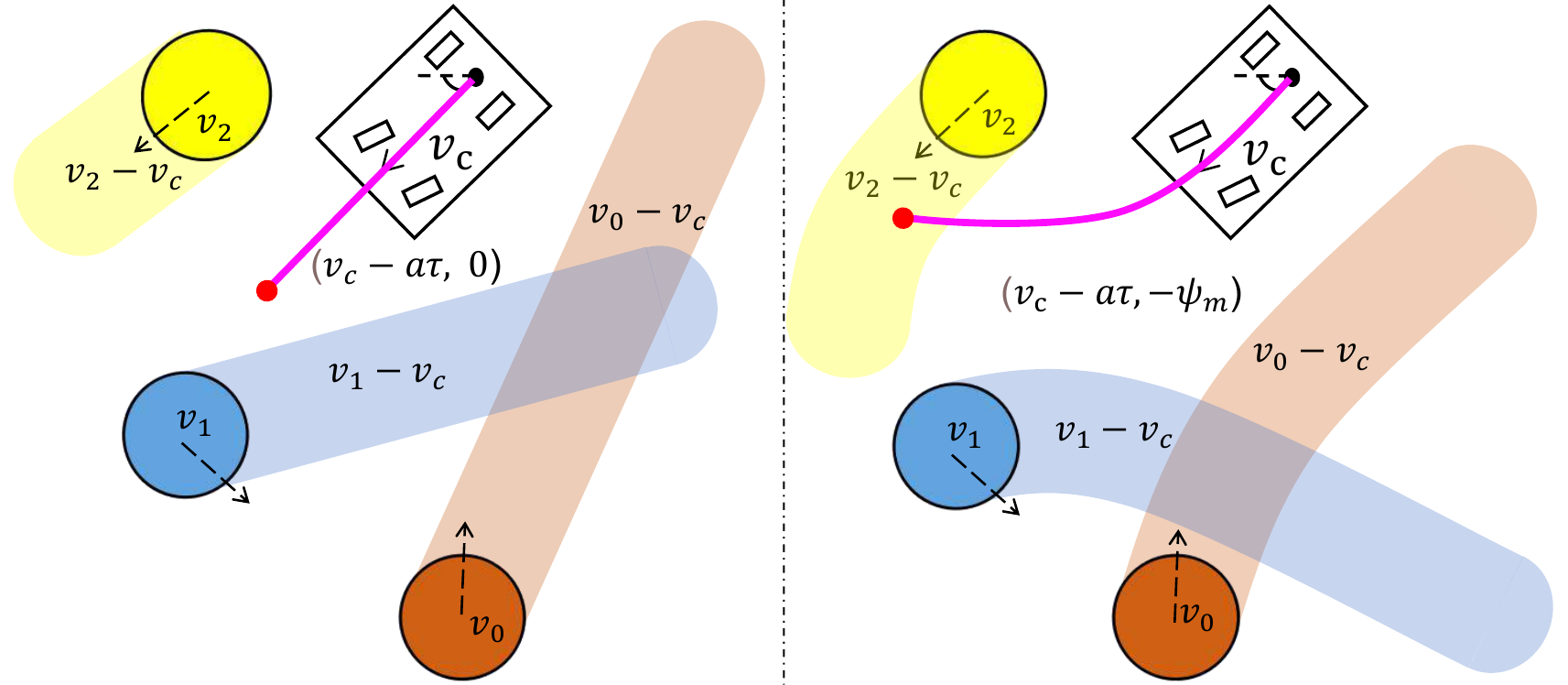}	
		\label{fig-ics}
	}
	\caption{Collision and ICS checking during graph searching.}
	\label{fig:collision}
	\vspace{-4mm}
\end{figure*}
This problem is typically solved by tree searching techniques, e.g., DFS and BFS, which however are not suitable for online trajectory generation due to their inefficiency. In this work, we propose to additionally discretize the state space for aggregating the motion primitives to enable online trajectory generation in highly dynamic environments. The details are presented in Alg. \ref{Generate-Succ} and \ref{astar}, which are used to generate motion primitives and perform graph searching, respectively. 
 
The graph is defined as $\mathcal{G}(\mathcal{X},\mathcal{U})$, where each vertex is a discretized grid cell corresponding to a robot state $\mathbf{x} \in \mathcal{X}$, and each edge is a motion primitive generated by the control input $\mathbf{u} \in \mathcal{U}_N \subset \mathcal{U}$. 
In Fig. \ref{special-case} (the $\phi$ dimension is not visualized), we demonstrate some cases of node expansion during graph building. It can be seen that there may exist multiple edges that connect two adjacent grid cells, and a grid cell may contain multiple state nodes. To make each grid being a vertex, we use the current optimal node $\mathbf{s}^*$ to represent that grid. In this way, the primitive tree is aggregated into a graph. 

In Alg. \ref{Generate-Succ}, we present the generation of collision-free motion primitives for expanding state nodes (see Section \ref{cc}), and the calculation of primitive cost and heuristic cost (see Section \ref{hd}). Based on this algorithm and the definition of $\mathcal{G}(\mathcal{X},\mathcal{U})$, we then perform online graph building and path searching with Alg. \ref{astar}, which is a variant of A* algorithm. The \textsc{HMAP} is a hash map used for storing state nodes $\mathbf{s}$. The map index of each node includes a hash key $\rm\mathbf{s}.key$ and a primitive index $\rm\mathbf{s}.idx$ (line 2), which are used to locate the grid cell that $\mathbf{s}$ belongs to and the position of $\mathbf{s}$ within the grid cell, respectively. In addition to the index field, $\mathbf{s}$ also records the robot state $\mathbf{x}$, i.e., $\rm\mathbf{s}.state$, the current cost value (line 3), and some other auxiliary fields. In lines 8-15, we check whether the goal or planning horizon is achieved, and make sure the local goal $\mathbf{s}_t$ is ICS free (see next section). Lines 18-42 present the node expansion and trajectory generation process. If a grid cell is never reached, we just expand a new state node for this cell (lines 38-40); Otherwise, we need to identify the optimal representation for this cell (lines 22-36). In Fig. \ref{special-case}, we visualize three cases (lines 26-28) that require to update the optimal representation, which is also the key implementation of aggregating motion primitives and reducing search efforts.

\subsection{Collision Checking}
\label{cc}

Collision checking is one of the most challenging and time-consuming problems in dynamic motion planning. The typical type of method is to discretize the target trajectory and then check collisions for each discretized waypoint by a forward simulation process. However, such a method cannot ensure the safety between waypoints, which may lead to severe collisions when there exist high-speed obstacles. To address this problem, we propose to calculate the minimum distance $d$ between motion primitive $\boldsymbol{\xi}_k$ and the obstacle trajectory for $t \in [k\tau,(k+1)\tau]$. If the condition $d > D$ is satisfied, where $D$ is the safety margin, the safety of $\boldsymbol{\xi}_k$ will be ensured. 

To achieve this point, as shown in Fig. \ref{fig-collision}, we first transform the obstacle trajectory by subtracting $\boldsymbol{\xi}_k$, which can be easily implemented by using Eq. \eqref{eq-solution}. Based on such a transformation, the problem is then converted to checking the minimum distance $d$ between the current position of the robot and the transformed trajectory of the obstacle. To accelerate the collision checking process, we only generate several waypoints (the red points in Fig. \ref{fig-collision}) and then connect these waypoints as line segments to linearize the transformed trajectory. As shown in the right side of Fig. \ref{fig-collision}, after the linearization, $d$ is then bounded within $[d^\prime-d_b, d^\prime+d_b]$, where $d^\prime$ is the minimum distance between the robot and the line segment. Therefore, the safety constraint in Eq. \eqref{optd} can be satisfied, if $d^\prime - d_b > D$ is ensured. Through the trajectory transformation and linearization, the collision checking is completed sufficiently and efficiently.

Another safety constraint is imposed on the local goal, i.e., $\mathbf{x}_t \not\in \mathcal{X}_{ICS}$. The ICS is defined as a state such that once the robot navigates into this state, it will inevitably collide with obstacles within a time interval. Identifying ICS is a time-consuming task, and thus in this work, we propose to test all the least-acceleration motion primitives starting from $\mathbf{x}_t$, demonstrated in Fig. \ref{fig-ics}, using our collision checking method. If there exists at least one collision-free primitive, it can be ensured that $\mathbf{x}_t \not\in \mathcal{X}_{ICS}$. Here, we only need to check the least-acceleration primitive for each steering angle (the most internal points in Fig. \ref{fig-primitive}), since if they are not ICS free, the outer primitives will not be ICS free.


\subsection{Heuristic Design}
\label{hd}

The length of Reed-Shepp curves without considering moving obstacles is adopted as the heuristic (line 6 in Alg. \ref{Generate-Succ}). These curves are dynamically feasible for  car kinematics model and have the shortest paths between the current state $\mathbf{s}$ and a given goal state $\mathbf{s}_g$. Reeds-Shepp heuristic is admissible in our problem, since it never overestimates the actual cost, i.e., $0 \leq h(\mathbf{s}, \mathbf{s}_g) \leq g(\mathbf{s}, \mathbf{s}_g)$. This heuristic is also consistent, which means that $h(\mathbf{s}_a, \mathbf{s}_c) \leq g(\mathbf{s}_a, \mathbf{s}_b)+h(\mathbf{s}_b, \mathbf{s}_c)$ holds for any $\mathbf{s}_a$ and $\mathbf{s}_b$. This can be easily proved by leveraging the facts that $h(\mathbf{s}_a, \mathbf{s}_c) \leq h(\mathbf{s}_a, \mathbf{s}_b)+h(\mathbf{s}_b, \mathbf{s}_c)$ and $0 \leq h(\mathbf{s}_a, \mathbf{s}_b) \leq g(\mathbf{s}_a, \mathbf{s}_b)$. In dynamic motion planning, the Reed-Shepp heuristic actually underestimates the actual cost too much, since the moving obstacles will significantly increase the path length. There exists a scale between the Reed-Shepp heuristic and the actual cost, i.e., $ g(\mathbf{s}, \mathbf{s}_g) = \alpha h(\mathbf{s}, \mathbf{s}_g), \alpha > 1$. A proper scale $\alpha$ on the heuristic can make it more informative, and enable the algorithm to converge quickly towards the solution. However, $\alpha$ depends on the future motions of moving obstacles and cannot be calculated at the current stage. In this work, we adopt an empirical value $\alpha = 1.3$, which can significantly improve the planning efficiency. The deficiency is that the solution given by Alg. \ref{astar} cannot be ensured optimal, since the scaled heuristic may become inadmissible sometimes.



\section{Experiments}
In this section, we first evaluate our method on a simulated environment to study its performance changes under different parameter settings. We then perform an overall evaluation on three benchmark datasets to compare our method with exiting work in the literature. Three baseline methods are adopted: the passive wait-and-go (WG) strategy, the classical velocity obstacle (VO) method \cite{gvo09}, and the up-to-date dynamic channel (DC) method \cite{dynamic-channel}. The experiment platform is a laptop with i5-9400 CPU 2.90 GHz with 8 GB of RAM. 
\begin{figure}[t]
	\centering
	\subfloat[The simulation environment.]{		
		\includegraphics[width=0.411\linewidth]{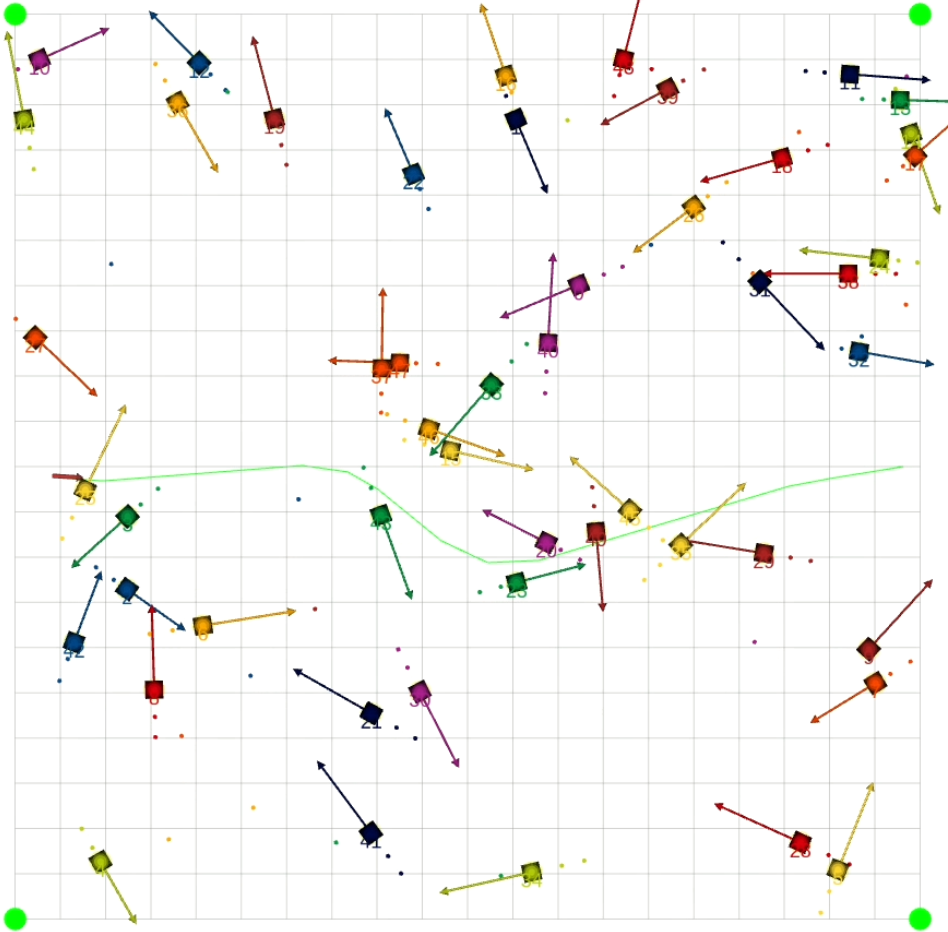}
		\label{simu}	
	}\hspace{2mm}
	\subfloat[Benchmark datasets \cite{eth-dataset, ucy-dataset}.]{		
		\includegraphics[width=0.43\linewidth]{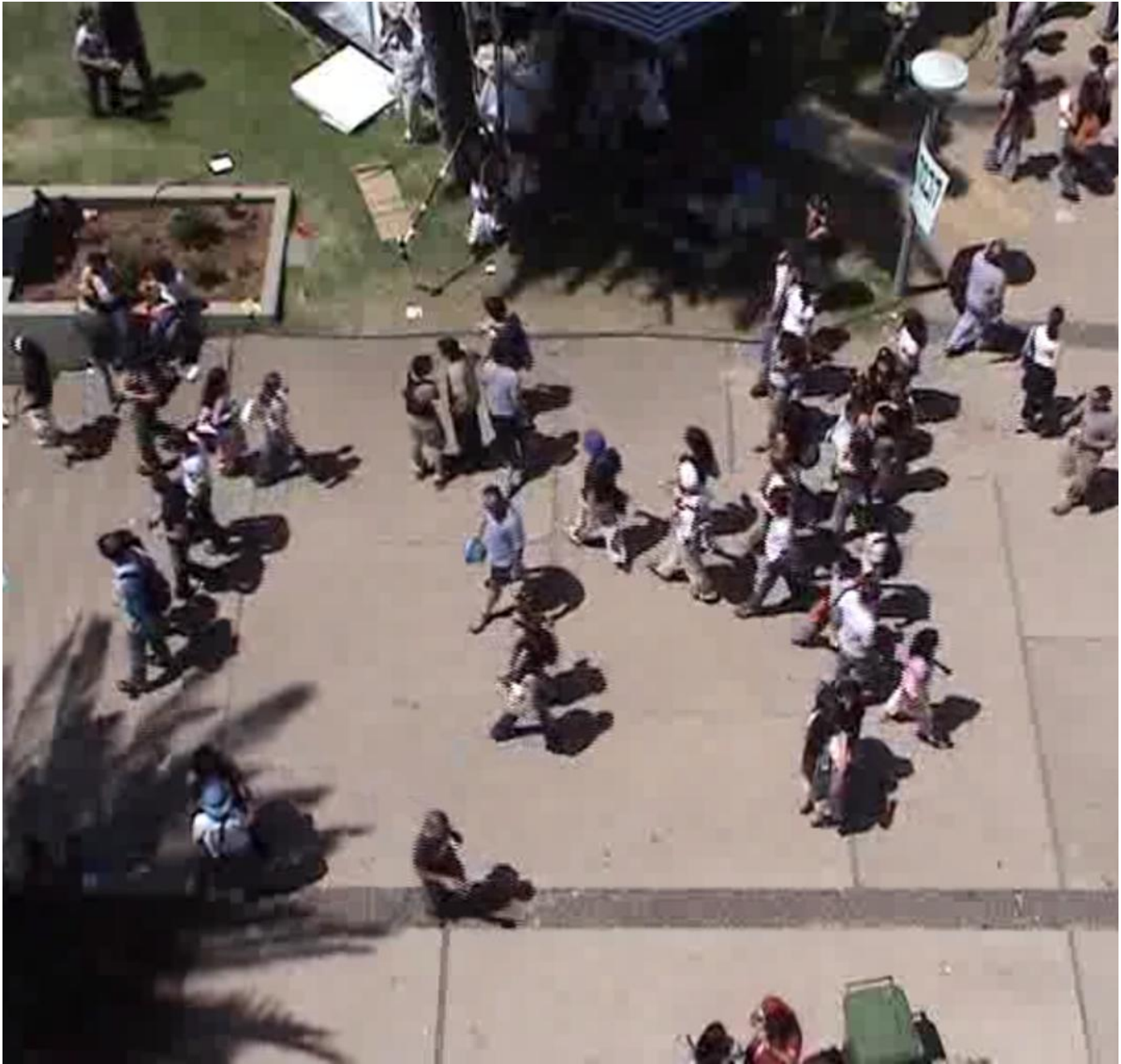}	
		\label{dataset}
	}
	\caption{The experiment environments and datasets used in this work.}
	\vspace{-4mm}
\end{figure}

\subsection{Experiment on Simulation Environment}
The performance of trajectory planning in dynamic environments is primarily influenced by three factors: the maximum velocity of the robot, the number of moving obstacles, and the safety margin. To quantify the influence, we conduct three control experiments in a 10m-by-10m simulation environment with multiple mobile agents. The moving direction of each agent is randomly initialized. The speed is generated by a uniform distribution from 1.2 to 2.0 m/s. If an agent moves outside the environment, a new agent will respawn. In this way, a constant obstacle density will be maintained. The current motion of the agents is available to the robot but with uncertainty on speed, which is modeled with a Gaussian noise $\mathcal{N}(0,0.1)$. The variance on speed can help test the robustness of different collision checking strategies. 

\begin{figure}[t]
	\centering
	\subfloat[Results under different number of agents.]{		
		\includegraphics[width=0.469\linewidth]{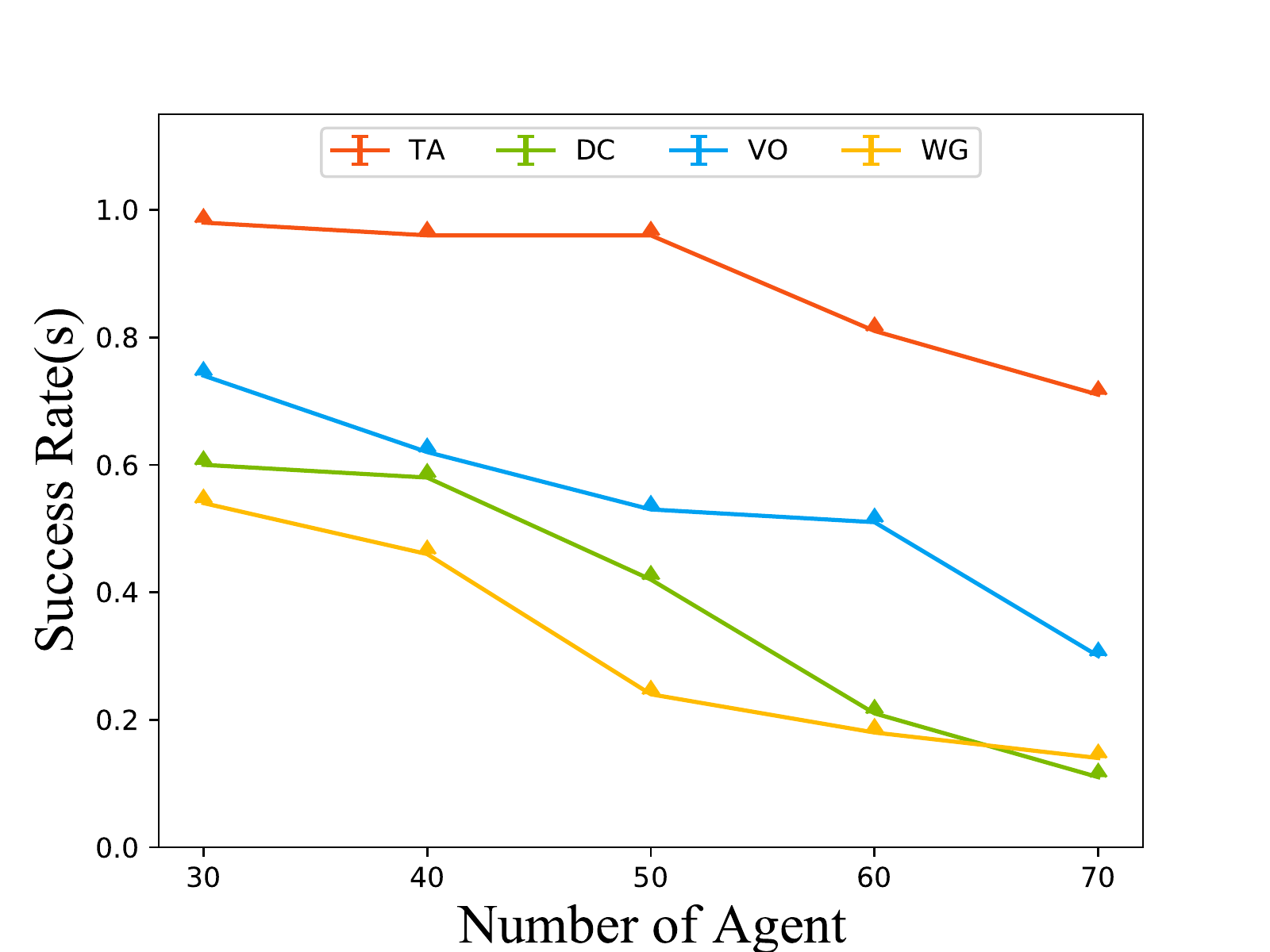}
		\includegraphics[width=0.469\linewidth]{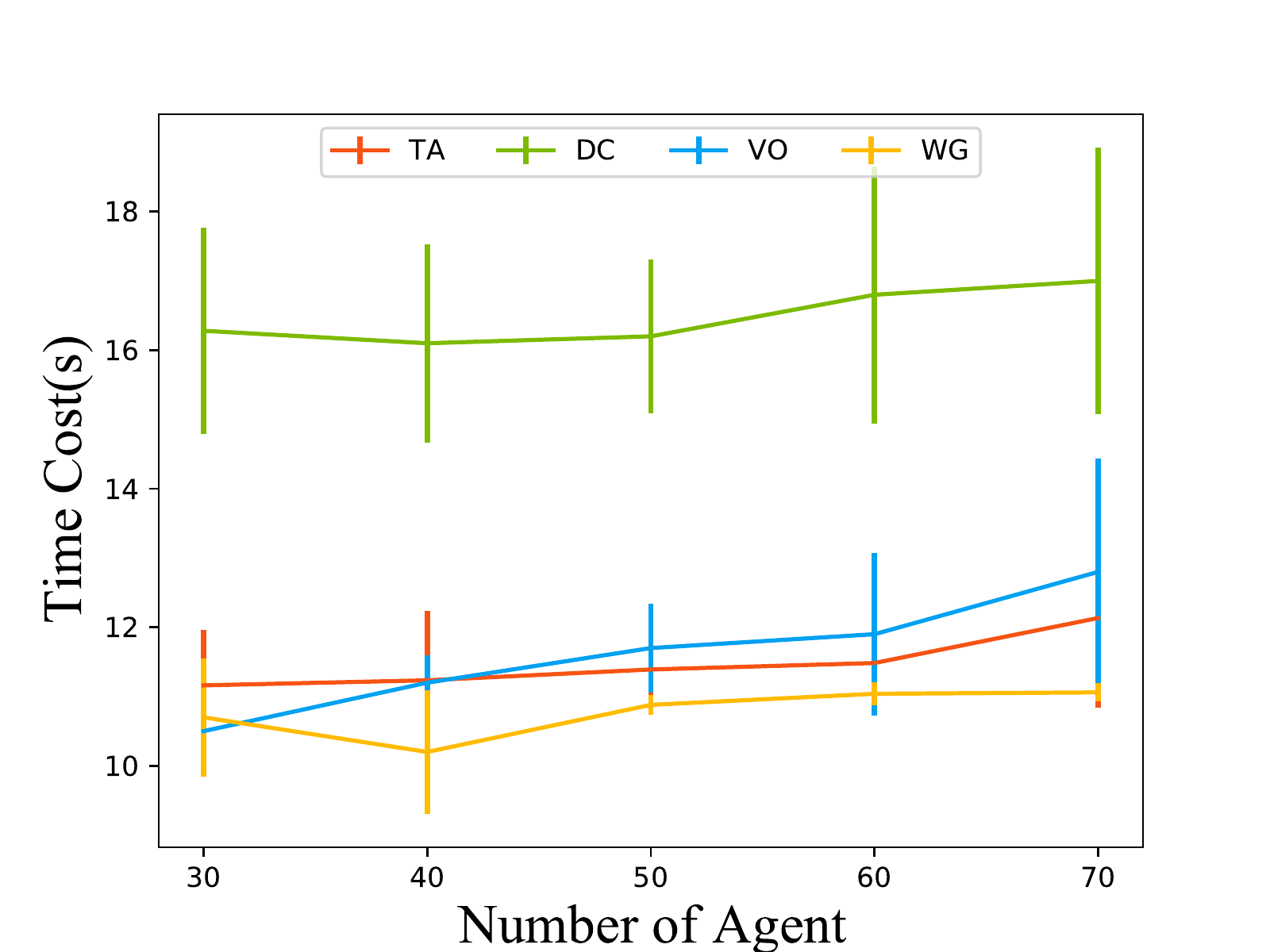}
		\label{sim-num}	
	}\vspace{-4mm}\\	
	\subfloat[Results under different safety margins.]{		
		\includegraphics[width=0.469\linewidth]{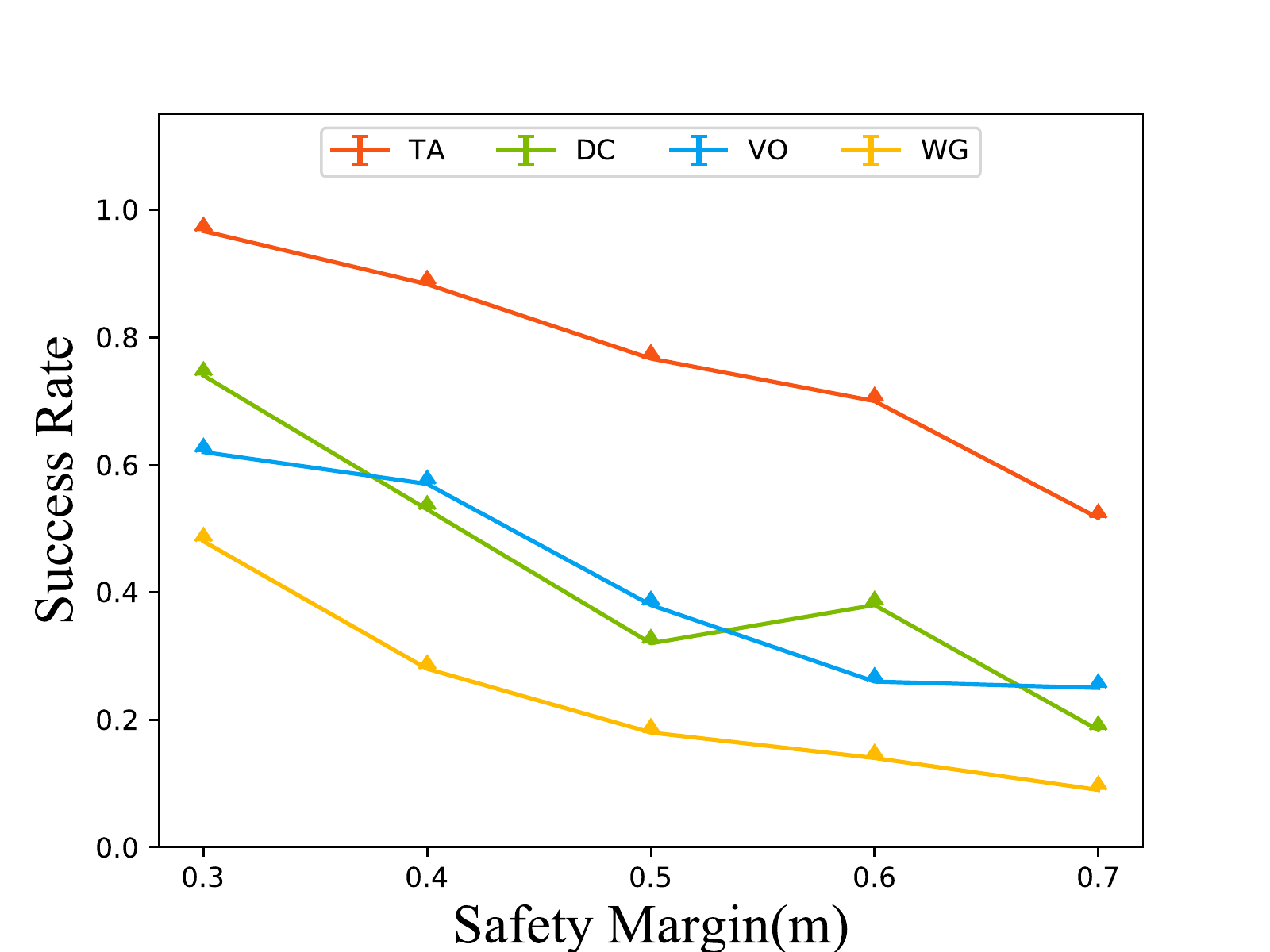}	
		\includegraphics[width=0.469\linewidth]{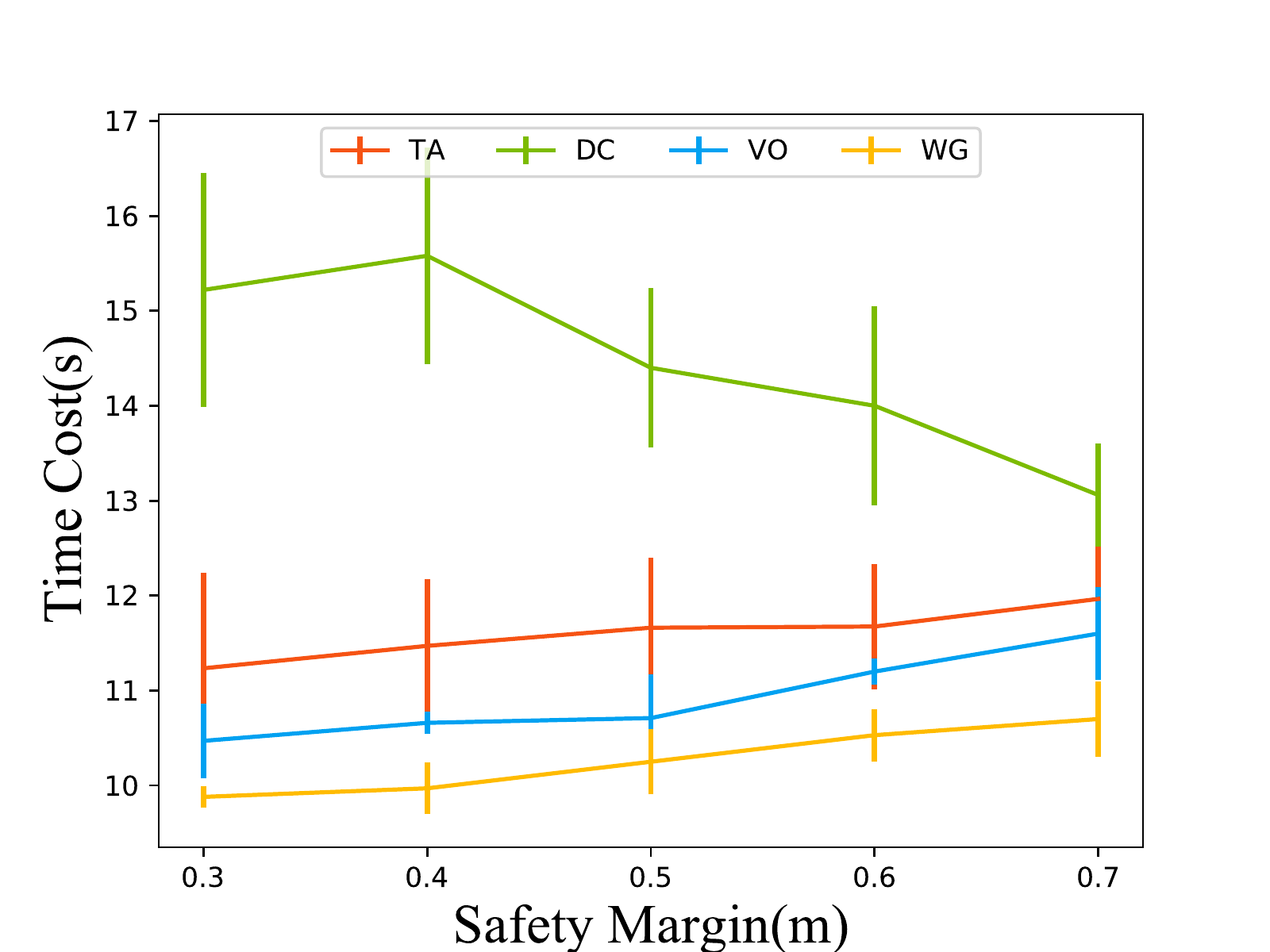}	
		\label{sim-dist}
	}
	
	\subfloat[Results under different maximum linear speeds.]{		
		\includegraphics[width=0.469\linewidth]{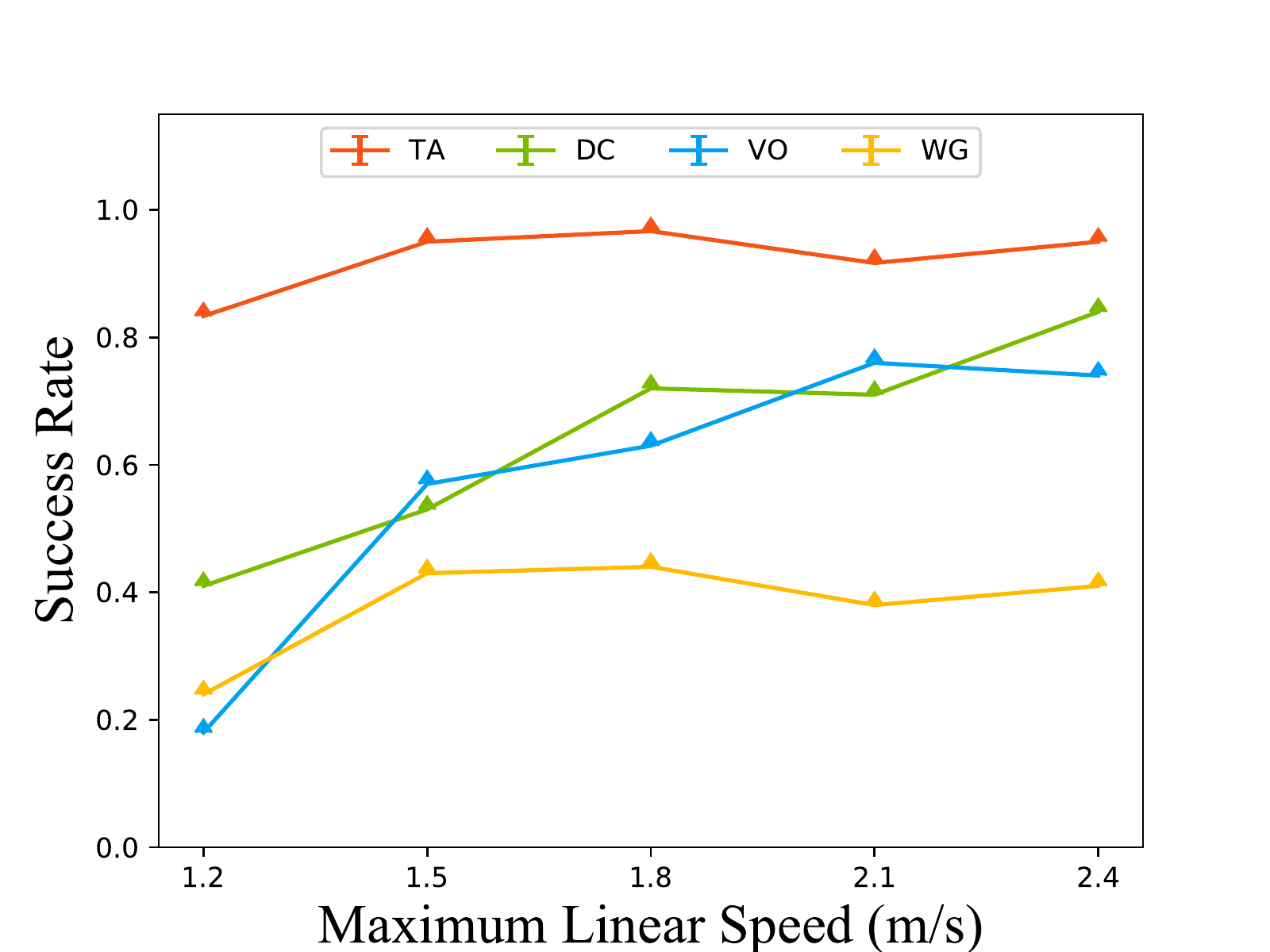}
		\includegraphics[width=0.469\linewidth]{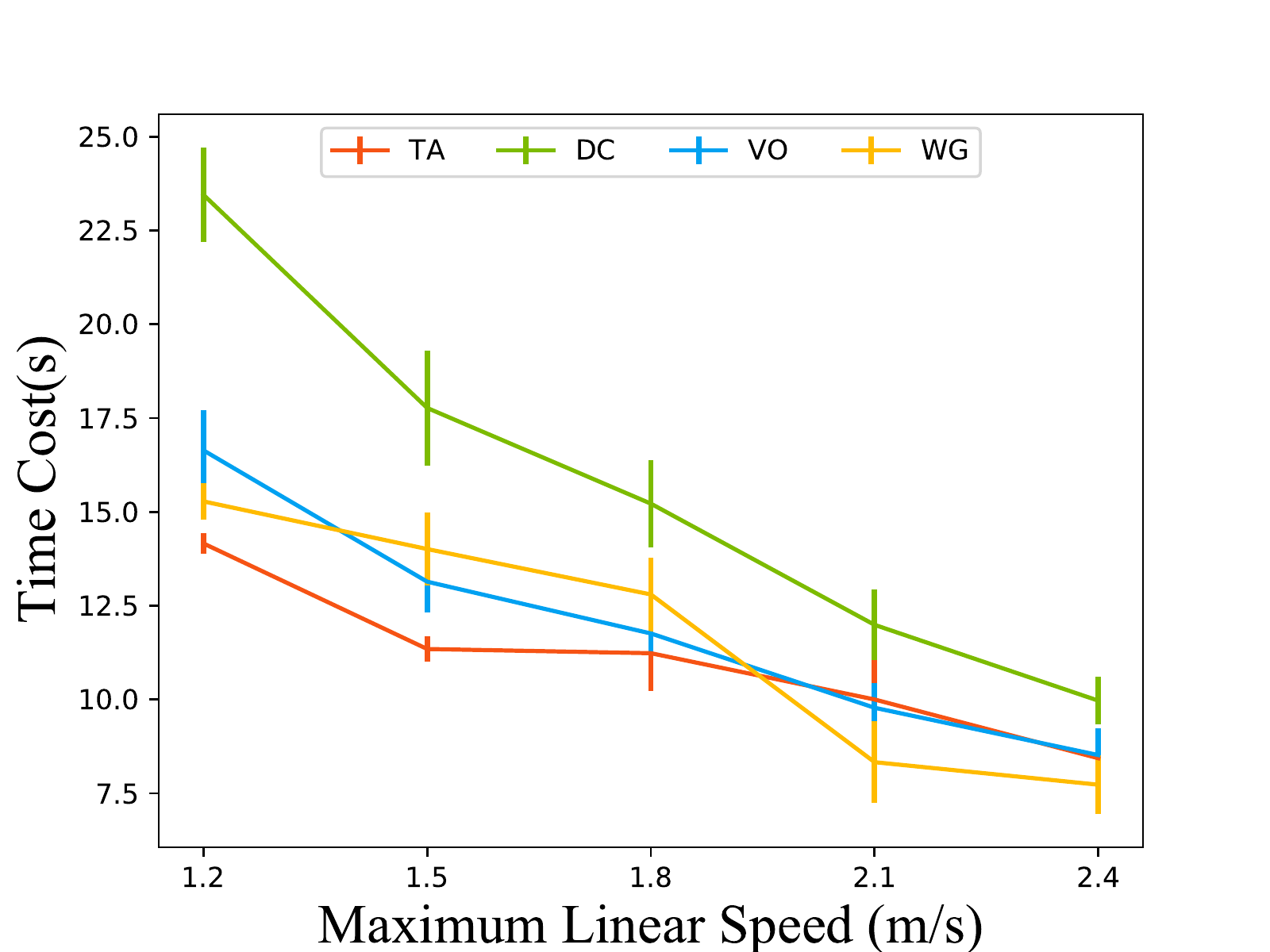}
		\label{sim-spd}
	}
	\caption{Performance on the simulation environment.}
	\label{sim-exp}
	\vspace{-4mm}
\end{figure}

The default parameter settings for the simulation are as follows: 40 moving agents, a safety margin of 0.3m, and a maximum linear speed of 1.8m/s for the robot. In each test of the experiment, one of the parameters is changed and the remaining is set to default. The middle points on the left and right sides of the environment, shown in Fig. \ref{simu}, are taken as planning start and goal, respectively. Each test repeats 30 times, and the average success rate and time cost are reported as evaluation metrics.
During experiments, a test is counted as a failure, if a collision happens or the robot cannot reach the goal in 30 seconds. The results are shown in Fig. \ref{sim-exp}. 

We can see that all planners exhibit a worse performance when the number of moving agents or the safe margin is increased, shown in Fig. \ref{sim-num} and \ref{sim-dist}. The reason is obvious that the feasible region in the planning space is occupied by the obstacles and margins, which makes it more difficult for the planner to find a feasible solution. Compared with the classical VO method, the latest DC planner shows a worse success rate in Fig. \ref{sim-num}, while maintaining a competitive performance in Fig. \ref{sim-dist}. Such a result is essentially caused by the different collision checking strategies. The DC method performs path searching on a sparse triangle graph generated by the Delaunay triangulation algorithm. The collision checking is limited to safely passing through a gate defined by the common edge of the current triangle with the adjacent one, which however cannot provide safety guarantee, especially in highly dynamic environments. The VO method considers all the surrounding obstacles for collision checking, thus exhibiting a better performance when the number of moving agents is increased. The proposed TA method can be regarded as a variant of VO enhanced by long-horizon planning, and thus shows the best performance among the four planners.

In terms of the maximum linear velocity in Fig. \ref{sim-spd}, we can see the success rate of DC is positively related to this factor, while the other three planners show a decreasing tendency after a certain test point. For the DC method, a higher speed can enable fast navigation through the edge of triangles under smaller topology changes of the graph, hence the success rate can be improved. For the TA and VO method, on the one hand, higher speeds increase the reactive ability of the robot and thus the success rate. On the other hand, higher speeds increase the horizon of collision checking, which is not beneficial for fast-planning. The success rate is thus affected to some extent, as shown in Fig. \ref{sim-spd}.

The simulation experiment shows that our proposed TA method can achieve the best performance on success rate and time cost, which significantly demonstrates its advantages over the baseline methods.

\begin{figure*}[t]
	\centering
	\subfloat[stu001]{
		\includegraphics[width=0.1945\textwidth]{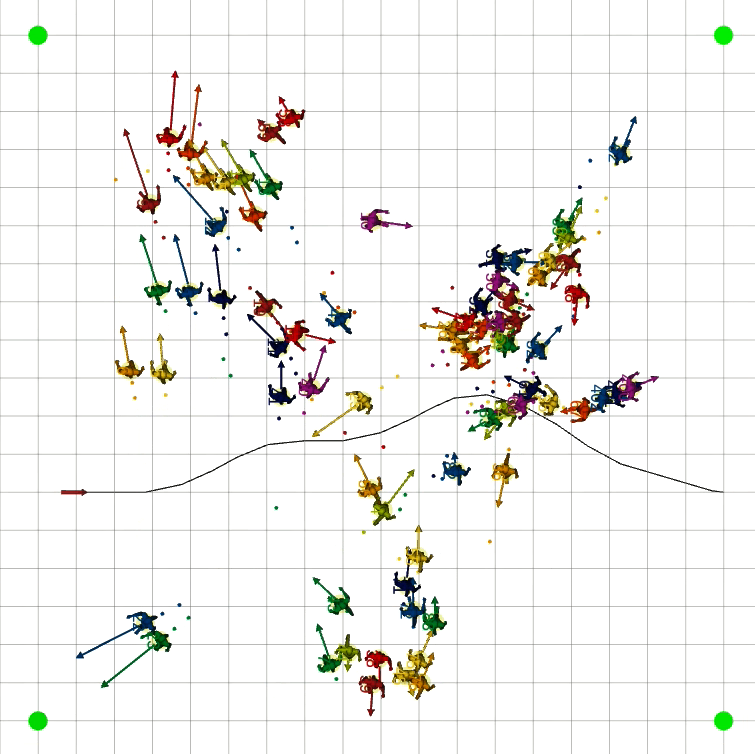}
	}\hspace{-3mm}
	\subfloat[stu003]{
		\includegraphics[width=0.194\textwidth]{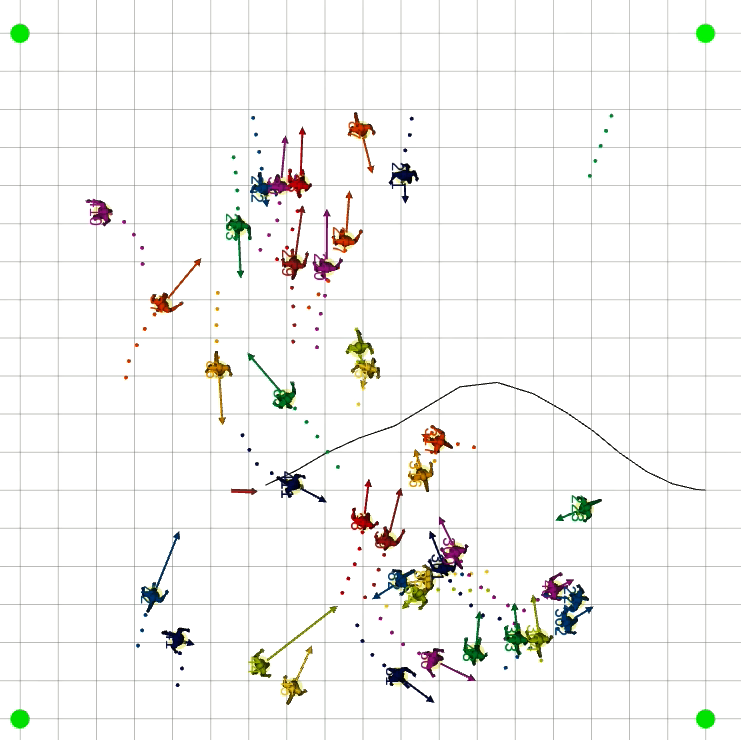}
	}\hspace{-3mm}
	\subfloat[zara01]{
		\includegraphics[width=0.195\textwidth]{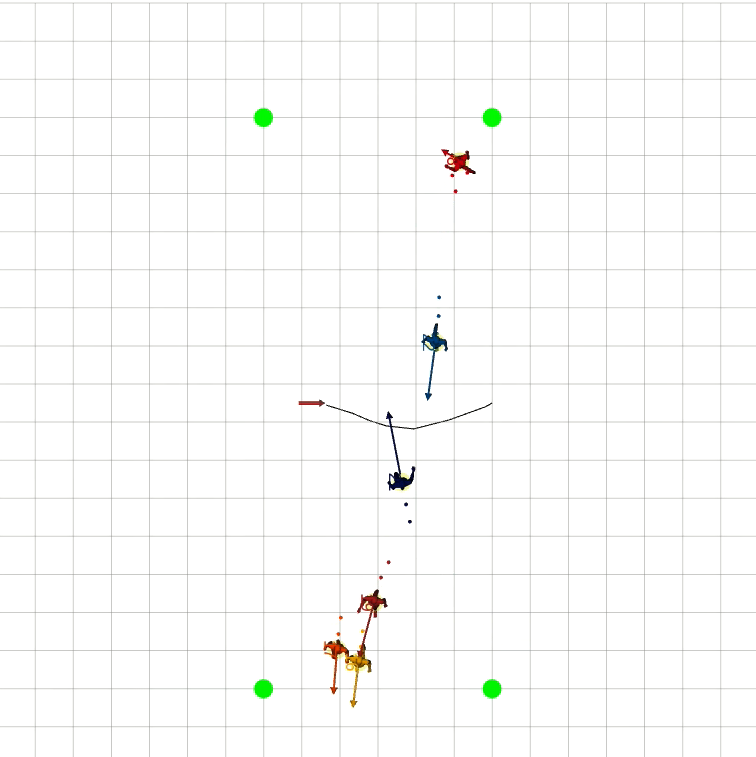}
	}\hspace{-3mm}
	\subfloat[zara03]{
		\includegraphics[width=0.193\textwidth]{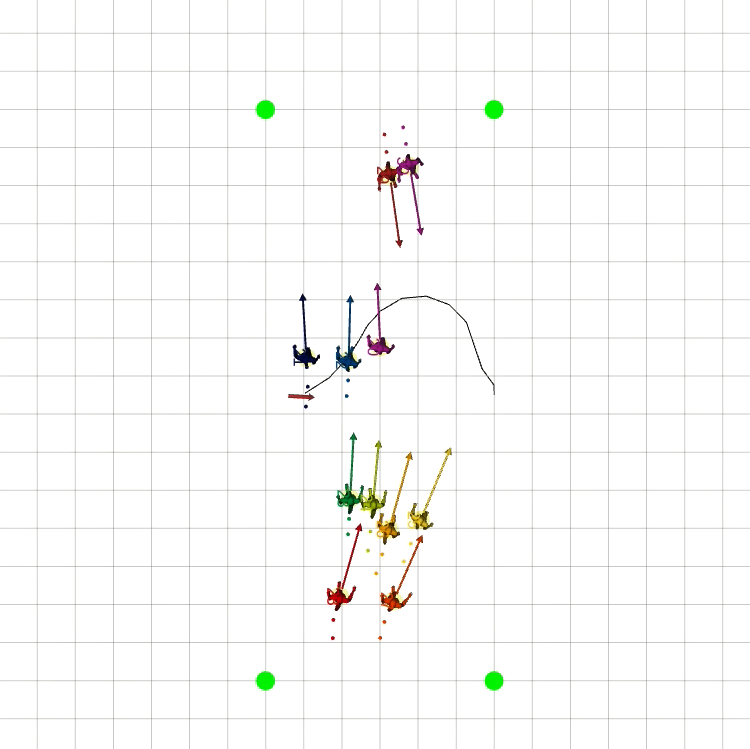}
	}\hspace{-3mm}
	\subfloat[biwi\_eth]{
		\includegraphics[width=0.1958\textwidth]{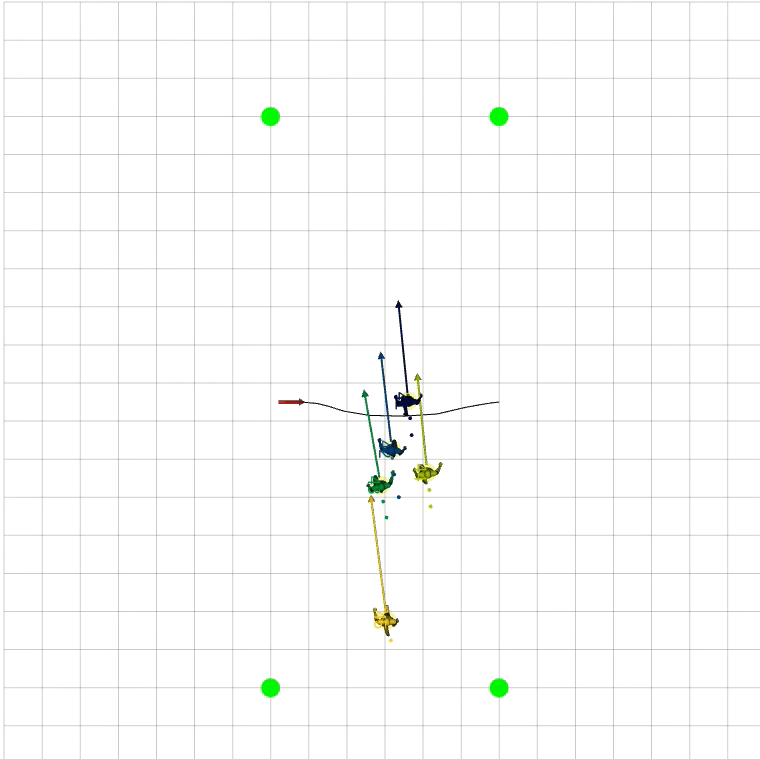}
	}   	
	\caption{The planning results of our online state-time planner on different data sequences.  }
	\label{fig:demo}
	\vspace{-6mm}
\end{figure*}
\begin{figure*}[t]
	\centering
	\subfloat[The results of success rate.]{		
		\includegraphics[width=0.33\linewidth]{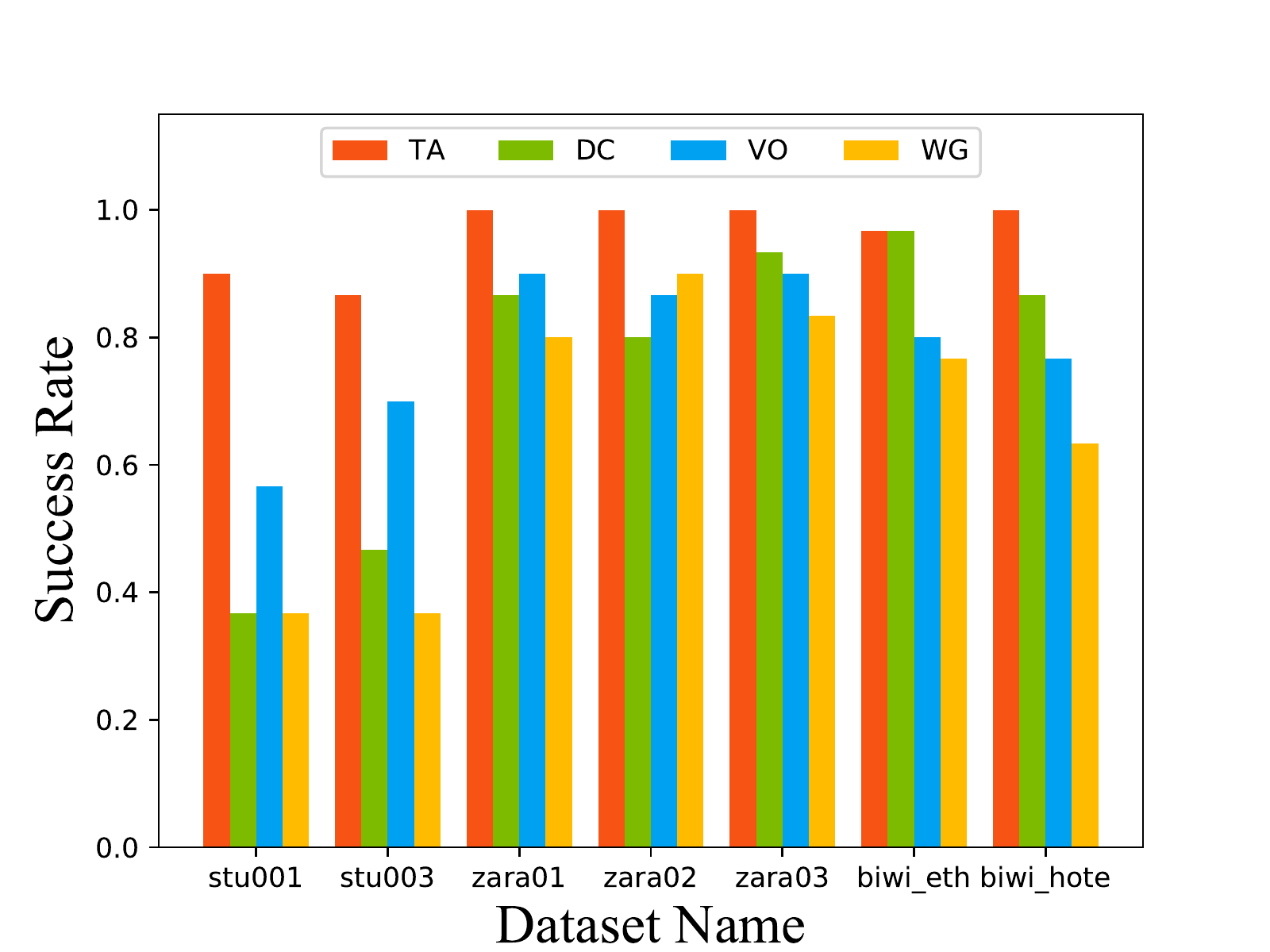}
		\label{fig-data_rate}	
	}
	\subfloat[The results average time cost.]{		
		\includegraphics[width=0.33\linewidth]{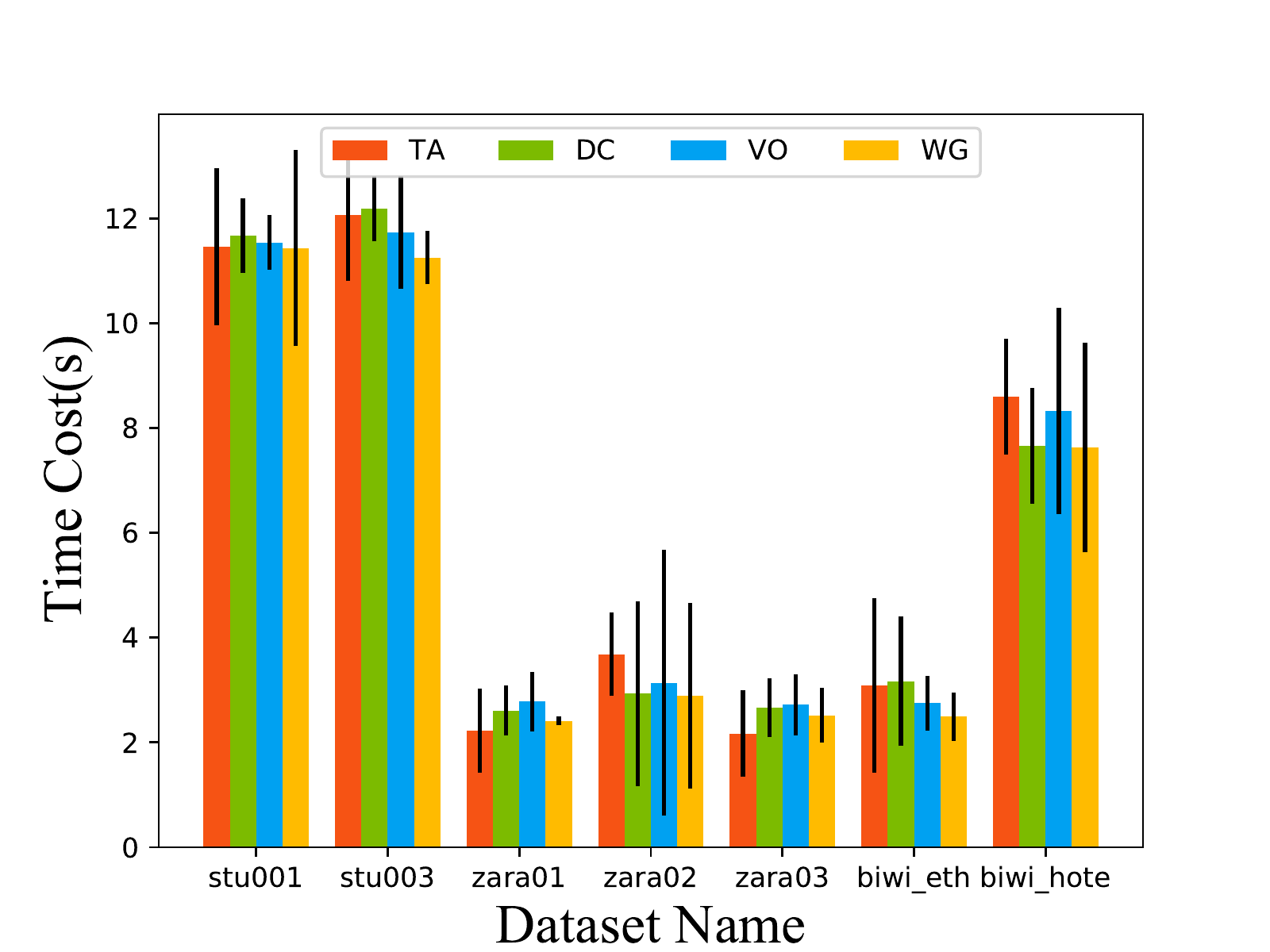}	
		\label{fig-data_t}
	}
	\subfloat[Speed and heading of robot]{		
		\includegraphics[width=0.33\linewidth]{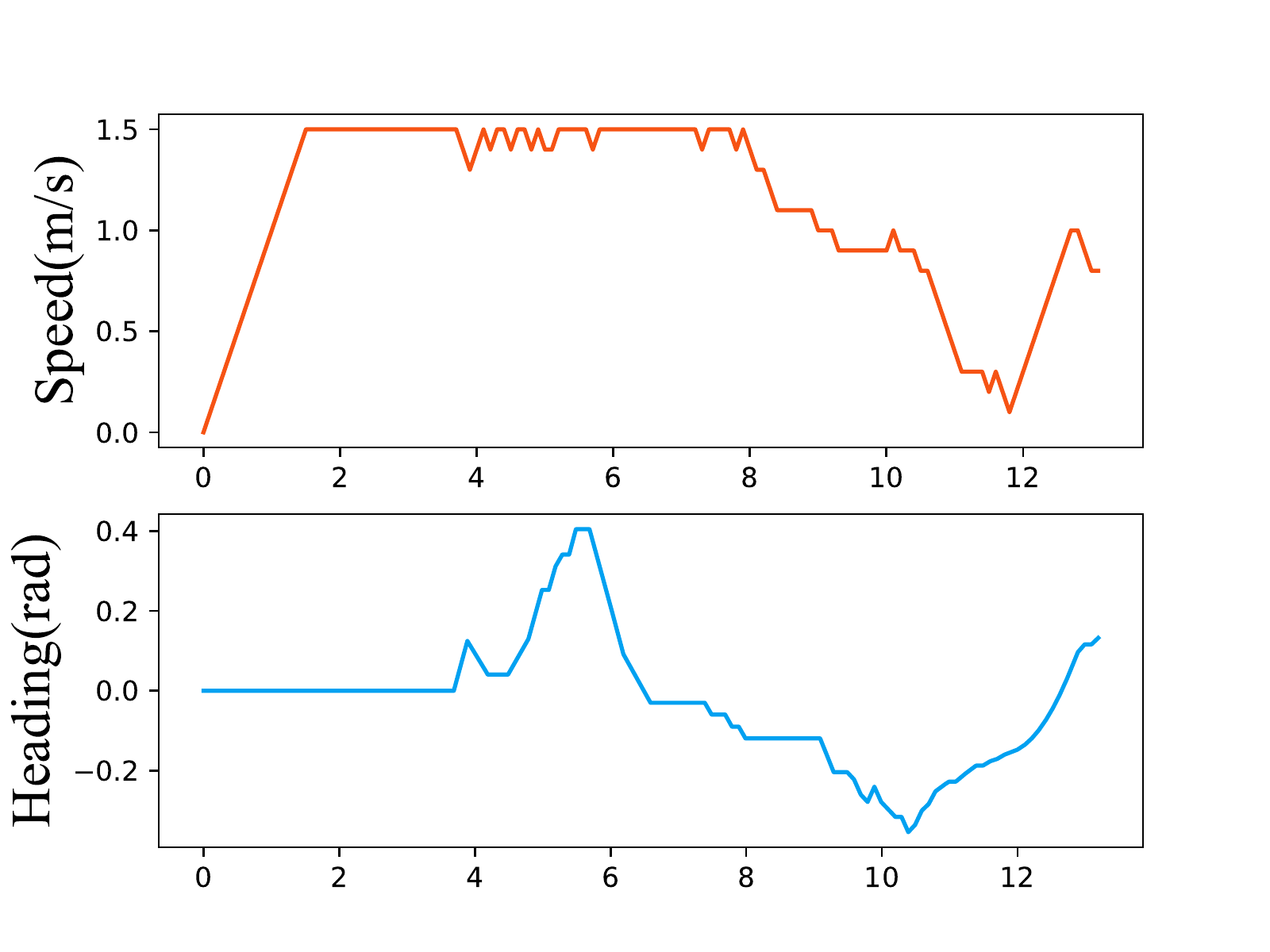}	
		\label{fig-car_state}
	}
	\caption{Performance on the benchmark datasets.}
	\label{fig:overall}
	\vspace{-4mm}
\end{figure*}

\subsection{Experiment on Benchmark Datasets}
We then evaluate our method on three public publicly available datasets from ETH [30] and UCY [31]. There are totally seven sequences of videos captured from different scenarios, i.e., \emph{stu001},  \emph{stu003},  \emph{zara01},  \emph{zara02},  \emph{zara03},  \emph{biwi\_eth}, and  \emph{biwi\_hotel}. All the sequences are interpolated and played at a frequency of 10Hz. As demonstrated in Fig. \ref{fig:demo}, the \emph{stu}, \emph{zara}, and \emph{biwi} sequences are with a high, medium, and low crowd density, respectively. During experiments, the middle points on the left and right sides of the environment, shown in Fig. \ref{fig:demo}, are taken as planning start and goal, respectively.  
The maximum linear speed is set to 1.5m/s. The safe distance is set to 0.4m. For each sequence, 30 tests are performed, and each test starts at a random time point of the sequence. The same evaluation metrics with the simulation experiment are adopted, and the experiment results are presented in Fig. \ref{fig:overall}.

In this part, we will analyze the success rate and time cost at the same time. WG achieves the lowest time cost in all datasets with a low success rate in \emph{stu001} and \emph{stu003}, as shown in Fig. 8a and 8b. On the one hand, WG always chooses the shortest path which produces the lowest time cost. On the other hand, the straight-line strategy has a large chance to collide in the environment with high crowd density, which results in a low success rate. As VO avoids the obstacle aggressively without considering the future motion of pedestrians, its performance on time cost is unstable. The fact that VO has time cost with large variance in \emph{zara01}, \emph{zara02} and \emph{biwi\_hotel} also support this statement. Time cost of TA is the highest in \emph{zara02} and \emph{biwi\_hotel}. The reasons are as follows: Firstly, TA can find a complex path toward the goal in the dense situation which contributes to a large average time cost. As shown in Fig 8a, TA still achieves a high success rate in \emph{zara02} and \emph{biwi\_hotel}. Secondly, TA may prefer a longer trajectory by going around the obstacle as a high penalty factor is given to deacceleration. We still claim that TA shows advantages over other planners. Because TA achieves the highest success rate in all the datasets and maintains a relatively low time cost. In general, we can observe a similar phenomenon in the simulation environment. As shown in Fig. \ref{fig-data_rate}, both TA and VO methods achieve a higher success rate in dense \emph{stu} sequences, while DC performs better than VO in sparse \emph{biwi} sequences.  \\
Fig. 8c shows the speed-time graph and heading-time graph of a sample trajectory generated by TA planner in dataset \emph{stu001}. We can observe that the speed and heading are smooth at the beginning, as TA prefers to go in a straight line without deacceleration. The rapid changes in linear speed are caused by unexpected collision avoidance, as the constant velocity assumption is inaccurate most of the time. The computation time of TA in different benchmark datasets is shown in Table \ref{table1}. In the experiment, we use 9 motion primitives for tree expansion and the planning horizon for each step is 0.5s. The coefficient for heuristic is $\alpha=1.3$. These are the key parameters related to the computation time. We can see that the proposed planner has a very high efficiency.

\begin{table}[htb]
	\centering
	\renewcommand{\arraystretch}{1.3}
	\caption{Computation time on benchmark dataset}
	\begin{center}
		
		\label{table1}
		
		\begin{tabular}{c|llllll}
			\hline
			Scene & \multicolumn{1}{c}{stu001} & \multicolumn{1}{c}{stu003} & \multicolumn{1}{c}{zara01} & \multicolumn{1}{c}{zara02}  & \multicolumn{1}{c}{biwi\_eth} & \multicolumn{1}{c}{biwi\_hotel} \\ \hline
			  mean(ms) &3.14& 4.77&3.14&3.64&1.85&3.12\\
			  std(ms)  &1.67& 2.61&2.12&2.50&0.98&1.25\\ 
			 \hline
		\end{tabular}
	\end{center}
	\label{time}
\end{table}

\section{Conclusions}
    
In this work, we propose a search-based partial motion planner to generate dynamically feasible trajectories for car-like robots in highly dynamic environments. The primitives generated by discretized control is over densely distributed in state-space, which is the major disadvantage of control space discretization method. We tackle this problem by selecting a set of optimal primitives for each discrete cell. We also propose a fast collision checking algorithm which linearizes relative motions between the robot and obstacles and then checks collisions by comparing a point-line distance. Benefiting from the efficient state representation and collision checking methods, the planner can generate feasible trajectories within milliseconds while maintaining a high success rate in highly dynamic environments. Extensive experiments also demonstrate its significant advantages. Our purposed method facilitates the trajectory planning for car-like robots based on control space discretization method in the highly dynamic environment. The major limitation of the algorithm lies in the un-informative heuristic functions, which are also the focus of our next-step work.

\bibliographystyle{IEEEtran}
\bibliography{ICRA2021}

\end{document}